\documentclass[preprint,12pt]{elsarticle}

\usepackage{amssymb}
\usepackage{amsmath}
\usepackage{graphicx}
\usepackage{subcaption}
\usepackage{booktabs}
\usepackage{multirow}

\usepackage{booktabs}   % for \toprule, \midrule, \bottomrule
\usepackage{pifont}     % for checkmarks and xmarks
\newcommand{\cmark}{\ding{51}} % checkmark
\newcommand{\xmark}{\ding{55}} % xmark

\usepackage[utf8]{inputenc}

\usepackage[bb=dsserif]{mathalpha}
\usepackage{bm}

\usepackage{color}
\usepackage{xurl}

\journal{Applied Soft Computing}

\begin{document}

\begin{frontmatter}

%% Title, authors and addresses

%% use the tnoteref command within \title for footnotes;
%% use the tnotetext command for theassociated footnote;
%% use the fnref command within \author or \affiliation for footnotes;
%% use the fntext command for theassociated footnote;
%% use the corref command within \author for corresponding author footnotes;
%% use the cortext command for theassociated footnote;
%% use the ead command for the email address,
%% and the form \ead[url] for the home page:
%% \title{Title\tnoteref{label1}}
%% \tnotetext[label1]{}
%% \author{Name\corref{cor1}\fnref{label2}}
%% \ead{email address}
%% \ead[url]{home page}
%% \fntext[label2]{}
%% \cortext[cor1]{}
%% \affiliation{organization={},
%%             addressline={},
%%             city={},
%%             postcode={},
%%             state={},
%%             country={}}
%% \fntext[label3]{}

\title{Leveraging graph neural networks and mobility data for COVID-19 forecasting}

\author[ppgccufop]{Fernando H. O. Duarte}
\author[decomufop]{Gladston J. P. Moreira}
\author[decomufop]{Eduardo J. S. Luz}
\author[cemaden]{Leonardo B. L. Santos}
\author[decomufop]{Vander L. S. Freitas\corref{cor1}}\ead{vander.freitas@ufop.edu.br}

\cortext[cor1]{Corresponding author}

%% Author affiliation
\affiliation[ppgccufop]{organization={Postgraduate Program in Computer Science, Federal University of Ouro Preto},%Department and Organization
            addressline={Morro do Cruzeiro}, 
            city={Ouro Preto},
            postcode={35402-163}, 
            state={MG},
            country={Brazil}}

\affiliation[decomufop]{organization={Department of Computing, Federal University of Ouro Preto},%Department and Organization
            addressline={Morro do Cruzeiro}, 
            city={Ouro Preto},
            postcode={35402-163}, 
            state={MG},
            country={Brazil}}

\affiliation[cemaden]{organization={National Center for Monitoring and Early Warning of Natural Disasters},
            addressline={Estrada Dr. Altino Bondensan, 500}, 
            city={Sao Jose dos Campos},
            postcode={12247-016}, 
            state={SP},
            country={Brazil}}

\begin{abstract}
The COVID-19 pandemic has claimed millions of lives, spurring the development of diverse forecasting models. In this context, the true utility of complex spatio-temporal architectures versus simpler temporal baselines remains a subject of debate. Here, we show that structural sparsification of the input graph and temporal granularity are determining factors for the effectiveness of Graph Neural Networks (GNNs). By leveraging human mobility networks in Brazil and China, we address a conflicting scenario in the literature: while standard LSTMs suffice for smooth, monotonic cumulative trends, GNNs outperform baselines when forecasting volatile daily case counts. We show that backbone extraction substantially enhances predictive stability and reduces predictive error by removing negligible connections. Our results indicate that incorporating spatial dependencies is essential for modeling complex dynamics. Specifically, GNN architectures such as GCRN and GCLSTM outperform the LSTM baseline (Nemenyi test, $p < 0.05$) on datasets from Brazil and China for daily case predictions. Lastly, we frame the problem as a binary classification task to better analyze the dependency between context sizes and prediction horizons.
\end{abstract}

% %%Graphical abstract
% \begin{graphicalabstract}
% %\includegraphics{grabs}
% \end{graphicalabstract}

% %%Research highlights
% \begin{highlights}
% \item Graph backbone extraction filters out negligible connections and enhances predictive stability.
% \item Comparing regression and classification tasks demonstrates that binary classification yields smoother, more interpretable results.
% \item Our prediction results present cross-regional consistency, considering the two case studies we analyze: Brazil and China.
% \item While standard LSTMs suffice for smooth, monotonic cumulative trends, GNNs better forecast volatile daily case counts.
% \end{highlights}

%% Keywords
\begin{keyword}
Graph Neural Networks \sep Time series forecasting \sep Mobility networks \sep Backbone extraction \sep COVID-19 
%% keywords here, in the form: keyword \sep keyword

%% PACS codes here, in the form: \PACS code \sep code

%% MSC codes here, in the form: \MSC code \sep code
%% or \MSC[2008] code \sep code (2000 is the default)

\end{keyword}

\end{frontmatter}

\section{Introduction}

The COVID-19 pandemic, caused by SARS-CoV-2, was declared a pandemic by the World Health Organization (WHO) on March 11, 2020. On November 26, 2025, more than 778 million confirmed cases and approximately 7.1 million deaths have been recorded globally. The first confirmed case in Brazil occurred on February 26, 2020, in São Paulo. Since then, the country has recorded more than 39 million confirmed cases and more than 716,000 deaths as of November 2025. In China, 96,203 cases and 4,636 deaths have been reported by the same date.

Droplets or aerosols transmit COVID-19, and these particles can remain suspended in the air, especially in closed and poorly ventilated environments. People's mobility plays a crucial role in the spread of the virus. Public transport, for example, can become a favorable environment for contagion due to crowding and proximity between individuals. Understanding mobility patterns is essential to predict \cite{Freitas_et_al_2020__CSP} and contain \cite{Freitas_et_al_2020__PeerJ} the spread of the disease.

Spatio-temporal models that combine mobility data with machine learning (ML) have gained significant attention for COVID-19 forecasting. Recent work highlights the effectiveness of incorporating spatial and temporal dependencies through graph-based approaches \cite{pathania2025leveraging}. In Banerjee et al. \cite{Banerjee_et_al_2022}, the authors represent graph vertices as locations, and the edges are weighted by geographical distance, applying an empirically derived threshold to construct the adjacency matrix. However, Brockmann and Helbing \cite{Brockmann_Helbing_2013} demonstrated that the effective distance, which is derived from mobility flows, predicts disease arrival times more accurately than geographical distance does. In this direction, \cite{MunozOrganero2023} proposed a Long Short-Term Memory (LSTM) model enhanced by mobility data from Madrid's bike-sharing system, achieving an 11.7\% improvement in accuracy over baseline models without mobility features. Similarly, Witzke et al. \cite{Witzke2023} demonstrated that Graph Neural Networks (GNNs) using mobility data improved trend forecasting during Germany's Omicron wave, particularly in identifying early change points where trends reversed direction.

Although the integration of mobility data improves accuracy, several limitations persist. Dense connectivity in mobility graphs often introduces noise, leading to less stable predictions, particularly in regression tasks. This issue is partially addressed in \cite{Kapoor2020} by a spatio-temporal GNN that incorporates both spatial and temporal edges. Still, the approach does not explore graph simplification techniques, such as backbone extraction, to filter out insignificant connections. Similarly, Sarkar et al. \cite{Sarkar2022} focused on explainable GNN frameworks for identifying high-risk regions but relied primarily on static population features, neglecting dynamic relationships observed in mobility networks.

Li et al. \cite{Li2024} applied machine learning techniques, including GNN and XGBoost, to analyze the spread of COVID-19 in Southern California, integrating socioeconomic characteristics and regional mobility data. While their approach demonstrated strong predictive performance, it relies on fixed temporal windows for weekly data, limiting its adaptability to dynamic changes in disease transmission patterns. This rigid structure fails to capture the evolving nature of epidemic trends. In contrast, the use of sliding windows with variable lengths introduces overlapping temporal sequences, enhancing the robustness of predictions by capturing both short- and long-term dependencies. Furthermore, exploring variable window sizes and prediction horizons remains underexplored within the COVID-19 forecasting task.

Duarte \textit{et al.} \cite{duarte2023} leverage Graph Convolutional Network (GCN) models integrated with recurrent neural networks (RNN) and LSTM networks to analyze spatio-temporal data in a regression task. The predictions are performed at the node level, where each node represents a city, and the GCN layer captures spatial information and vehicle flows between them. The study employed a fixed window size of 14 days to predict the number of cumulative cases for the next day. The results indicated a trade-off: while the GNN-based models, specifically GCRN (GCN+RNN) \cite{gcrn} and GCLSTM (GCN+LSTM) \cite{gclstm}, demonstrated greater stability, evidenced by lower standard deviation and significantly lower maximum errors, the baseline models (LSTM and Prophet) achieved better average and minimum error metrics. However, this previous work has limitations regarding fixed contexts and prediction horizons. Furthermore, the focus on cumulative cases proved to be a relatively simple regression task that standard LSTMs can solve effectively without leveraging the complex topological structure of the input graph, in contrast to the more volatile daily-variation data explored herein.

Despite advances in ML-based forecasting, most studies formulate COVID-19 prediction as a regression task~\cite{duarte2023,COMITO2022102286,JOSHI2020104502}. However, reframing the problem as a classification task \cite{su12062427} can provide significant benefits, including simpler results, smoother predictions, and enhanced interpretability, particularly when categorizing outcomes into meaningful states, such as ``stable'' and ``alert''. Furthermore, comparative analyses across multiple regional datasets remain limited despite their potential to uncover valuable insights into regional variations and shared patterns in disease dynamics. Moreover, the predominant focus on studies in a single region restricts the generalizability and broader applicability of the findings, highlighting the need for research in diverse geographic contexts~\cite{Freitas_et_al_2020__CSP,Sarkar2022,Li2024,duarte2023,su12062427,JOSHI2020104502,Duarte2023363,PredictingGraphNN,xie2022visualization,SpreadCovid,review1}.

From these identified gaps emerge several hypotheses. We posit that extracting the backbone \cite{Menczer_Fortunato_Davis_2020,Ferreira_et_al_2022__backbone,Serrano_2009_disparity_filter} of the mobility network can enhance accuracy by highlighting essential spatial relationships; that using sliding windows with overlapping sequences surpasses fixed-window approaches by capturing a richer temporal structure; that formulating the prediction task as binary classification (stable vs. alert states) produces more stable and interpretable results compared to regression; that forecasting daily cases instead of accumulated counts presents a significantly more challenging task due to the non-well-behaved nature of the time series, thereby making the topological information from mobility data crucial for prediction; that the consistent performance patterns across prediction horizons and window sizes can be observed in datasets from both Brazil and China, demonstrating robustness to geographic variation; and that the GCRN and GCLSTM models perform comparably, validating the reliability of graph-based spatio-temporal forecasting methods in diverse contexts. 

This work extends the study by Duarte \textit{et al.} \cite{duarte2023} by addressing these gaps and incorporating the above-mentioned hypotheses into a comprehensive experimental framework. Our main contributions include: (i) establishing topological sparsification (via backbone extraction) as a critical structural prior for spatio-temporal forecasting. We show that raw mobility networks introduce a low signal-to-noise ratio that degrades GNN performance, whereas the backbone preserves the essential causal pathways of transmission; (ii) demonstrating a fundamental distinction in model applicability: while standard LSTMs are sufficient for the smooth, monotonic nature of cumulative cases, GNNs provide a superior inductive bias for modeling the volatile, stochastic dynamics of daily case variations; (iii) implementing sliding windows of variable size and prediction horizons to capture temporal dependencies dynamically; (iv) framing the prediction as a binary classification task to yield smoother, more interpretable results for decision-making; and (v) validating the regional consistency and robustness of the proposed framework through a comparative analysis across distinct epidemiological contexts in Brazil and China.

Our results indicate that GNN-based approaches are comparable to the LSTM baseline when the time series represent accumulated disease cases, as these series tend to be well-behaved. However, when forecasting daily cases, the mobility network plays a pivotal role, making predictions significantly more accurate, as measured by Root Mean Square Error (RMSE), by capturing the spatial dynamics of the spread. 

The remainder of this paper is organized as follows: Section \ref{sec:materials_methods} provides an overview of the materials and methods, including key definitions, datasets, and details of the employed methodology. Section \ref{sec:results} presents and discusses the results in the context of existing literature. Finally, Section \ref{sec:conclusions} summarizes the findings and concludes the study.

\section{Materials and methods}
\label{sec:materials_methods}

\subsection{Mathematical preliminaries}

A graph $G(V,E)$ has a set of vertices (nodes) $V = \{v_1, v_2, \cdots, v_N \}$ and edges (links) $E$, totaling $N=|V|$ nodes and $L=|E|$ edges. Mathematically, the adjacency matrix $\textbf{A} \in \mathbb{R}^{N \times N}$ provides the corresponding wiring diagram, with $\textbf{A}_{ij}=1$ indicating when nodes $i$ and $j$ are connected, or zero otherwise, for unweighted graphs. When it comes to weighted graphs, as in our case with mobility networks, $\textbf{A}_{ij}$ might take different values to describe connections. The $\textbf{D}_{ii} = \sum_j \textbf{A}_{ij}$ is the degree matrix, with zeros outside the main diagonal, and $\textbf{X} \in \mathbb{R}^{N \times C}$ is the feature vector with $C$ dimensions for each vertex, i.e., the input data for training the models. 

\subsection{Network Backbone extraction}
\label{sec:backbone}

A node might have connections that are not statistically relevant, which could potentially deteriorate the results of regression and classification tasks. Furthermore, these superfluous connections lead to computational graphs that are unnecessarily large, ultimately contributing to the oversmoothing phenomenon \cite{Rusch_et_al_2023_oversmoothing}. One possibility to address this is to employ the so-called Disparity Filter \cite{Serrano_2009_disparity_filter,Menczer_Fortunato_Davis_2020}, which contrasts such connections with the random expectation: 
\begin{equation}
    p_{ij} = \left ( 1 - \frac{\textbf{A}_{ij}}{s_{i}} \right )^{k_i - 1},
\end{equation}
in which $k_i$ and $s_i$ are the degree and strength of node $i$, respectively, and $\textbf{A}_{ij}$ is the weight of link $ij$. The degree $k_i = \sum_{j=1}^N \theta(\textbf{A}_{ij})$ denotes the number of connections node $i$ has, with $\theta$ being the heaviside step function. Similarly, the node strength $s_i = \sum_{j=1}^N \textbf{A}_{ij}$ is the weight accumulated by neighboring nodes. 

The link is preserved if $p_{ij}$ is below a desired significance level $\alpha$. Otherwise, it is removed. Thus, the lower the $\alpha$, the sparser the network. In addition, one has to consider that each link has two ends, $i$ and $j$, which requires choosing one via a criterion such as the largest or smallest. In this work, we pick the smallest, which leads to sparser networks.

\subsection{Graph Convolutional Networks}

In recent years, graph neural networks (GNNs) have emerged as a powerful tool for handling graph-structured data in classification and regression problems at the node, edge, and graph levels \cite{Wu2021}. GNNs are categorized into several types, such as recurrent (RecGNNs), convolutional (GCNs), autoencoders (GAEs), and spatio-temporal (STGNNs). 

GCNs are a specific type that adapts the convolutional task, well-known in the Euclidean domain for images, to the non-Euclidean domain, where the notion of neighborhood operates at the graph level \cite{Niepert}. Each node has a computational network, a subgraph that aggregates its neighborhood up to a chosen depth. As nodes with many neighbors propagate more quickly than more isolated nodes \cite{kipf2016semi}, one assigns greater weights to the features of nodes with fewer neighbors, thus balancing the influences. 

Concerning the node $v$, the first layer receives its features $x_v$, so the corresponding activation is $h_v^0 = x_v$. Then, for every consecutive layer, one combines data from its neighbors $N(v)$, sums it with its current activation, and does it recursively until reaching the final layer $k$, where the node embeddings $z^{(v)}$ can be retrieved:
\begin{align}
\left \{
\begin{aligned}
h_v^0 &= x_v, \\
h_v^k &= \sigma \left[ W_k \sum_{u \in \mathcal{N}(v)} \left( \frac{h_u^{k-1}}{|\mathcal{N}(v)|} \right) + B_K h_v^{k-1} \right], \\
z^{(v)} &= h_v^k,
\end{aligned}
\right.
\label{eq:gcn_activations}
\end{align}
\noindent so that $B_k$ is the weight associated with the node itself, $W_k$ is the weight matrix of the layer $k$, and $\sigma$ is the activation function. For the matrix form of Eq. \eqref{eq:gcn_activations}, let $\widehat{A} = A + I_N$, $A$ be the weighted adjacency matrix of the subgraph, $I_N$ be the identity matrix of order $N$, and $\widehat{D}$ be the diagonal degree matrix of $\widehat{A}$. The output of the convolution layer $k+1$ is as follows: 
\begin{equation}
    H^{k+1} = \sigma (\widehat{D}^{-1/2}\widehat{A}\widehat{D}^{-1/2}H^{k}W^{k+1}).
    \label{eq:H}
\end{equation}

\subsection{Time series prediction}

% To predict a future sequence of length $F$ (prediction horizon), one feeds the model with a sequence of length $l$ (window size) of observed data at previous timestamps:
% \begin{equation}
%      \hat{x}_{t+1}, ... , \hat{x}_{t+F} = \max_{x_{t+1}, ... , x_{t+F} } P(x_{t+1}, ... , x_{t+F} | x_{t-l+1}, ... , x_{t}),
% \end{equation}
% \noindent where $x_{t}$ is a snapshot at time $t$ of the features and $\hat{x}_{t}$ is the prediction.

% In this setup, $x_t$ represents the time series value in time $t$ between all nodes within a static, directed, and weighted graph. In contrast, a snapshot $X$ is a matrix $X \in \mathbb{R}^{N \times C}$, which captures data from a time series at $C$ time points.

Let $G = (V, E)$ denote a weighted graph with $N$ nodes. At each time step $t$, a graph signal is defined as $X_t \in \mathbb{R}^{N \times C}$, where each node is associated with a $C$-dimensional feature vector.

Given a sequence of past observations of length $l$ (window size), the goal is to predict a future sequence of length $F$ (prediction horizon). This can be formulated as learning a function $f_\theta$ such that:
\begin{equation}
    (\hat{X}_{t+1}, \ldots, \hat{X}_{t+F}) = f_\theta(X_{t-l+1}, \ldots, X_t; G),
\end{equation}
\noindent where $f_\theta$ is parameterized by a graph recurrent neural network, such as GCLSTM or GCRNN, and $\hat{X}_{t+k}$ denotes the predicted graph signal at time $t+k$.

\subsection{GNNs for time series prediction}

\begin{figure}[htpb!]
    \centering
    \includegraphics[width=0.9\textwidth]{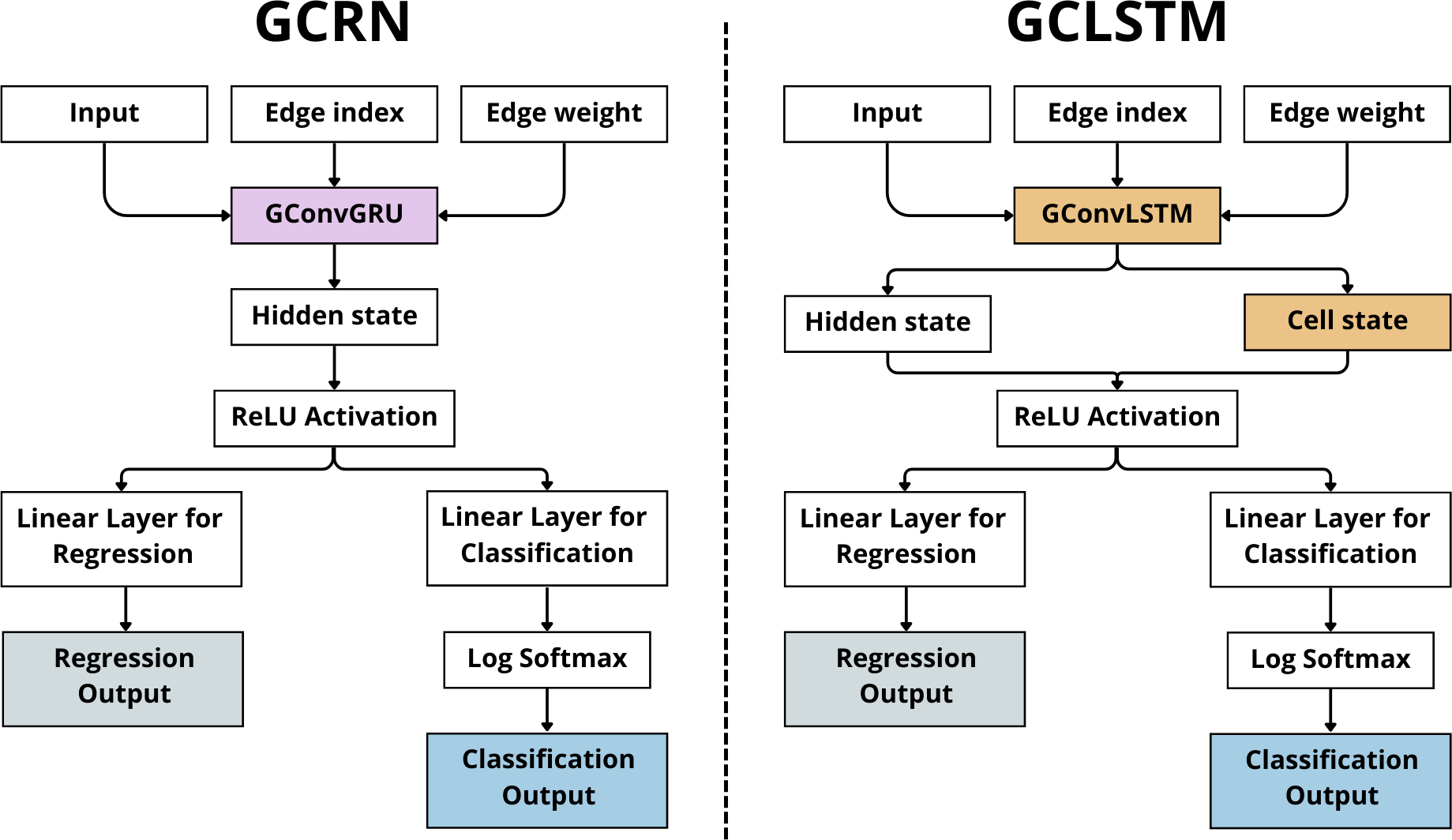}
    \caption{Simplified representation of the GCRN and GCLSTM models.}
    \label{fig:models}
\end{figure}

Graph Convolutional Recurrent Networks (GCRN) architectures \cite{gcrn} and Graph Convolutional LSTM (GCLSTM) \cite{gclstm} (Fig. \ref{fig:models}) are Graph Neural Networks (GNNs) that deal with sequential data. The GCRN combines graph convolutions with temporal recurrence, while the GCLSTM employs LSTM units to model temporal dynamics on graphs. Each architecture is adapted to support regression and classification tasks, differing only in the output layer: \textit{Linear} for regression and \textit{Softmax} for classification. 

% BACKUP of the last Figure
% \begin{figure}[htpb!]
%     \centering
%     \begin{subfigure}[b]{0.49\textwidth}
%         \includegraphics[width=\textwidth]{GCRN-2024-12-02-215238.png}
%         \caption{GCRN model architecture.}
%         \label{fig:GCRN}
%     \end{subfigure}
%     \hfill
%     \begin{subfigure}[b]{0.49\textwidth}
%         \includegraphics[width=\textwidth]{GCLSTM-2024-12-02-215509.png}
%         \caption{GCLSTM model architecture.}
%         \label{fig:GCLSTM}
%     \end{subfigure}
%     \caption{Simplified representation of the GCRN and GCLSTM models.}
%     \label{fig:models}
% \end{figure}

The GConvLSTM (Graph Convolutional Long Short-Term Memory) and GConvGRU (Graph Convolutional Gated Recurrent Unit) layers are extensions of classical recurrent architectures (LSTM and GRU, respectively) for graph-structured data. Both combine the ability to model temporal sequences with the ability to capture spatial relationships. The GCLSTM architecture uses GConvLSTM, while the GCRN incorporates GConvGRU, as shown in Figure \ref{fig:models}. 

The GConvLSTM layer uses graph convolutions to process sequences of data on graphs. Its main features include:
\begin{itemize}
    \item Input and output: Receives node data, edge indices, and optionally edge weights. Produces hidden and cell states.
    \item Graph Convolution: Utilizes graph convolution to capture interactions between the graph's nodes.
    \item Laplacian Normalization: Employs different normalization schemes to control information propagation.
    \item Gate Computation: Computes input, forget, cell, and output gates using graph convolutions and standard LSTM operations.
    \item State Updates: Updates hidden and cell states by integrating graph information.
\end{itemize}

The GConvGRU layer, in turn, is based on the GRU architecture and offers a simpler alternative to GConvLSTM. Its main features include:
\begin{itemize}
\item Structure: Similar to GConvLSTM, but with fewer parameters.
\item Gates: Uses update and reset gates.
\item Gate Computation: Employs graph convolutions to compute the gates.
\item State Update: Updates the hidden state more directly.
\end{itemize}

\subsection{Datasets}

\subsubsection{Brazilian mobility network}

The Brazilian Institute of Geography and Statistics (IBGE) \cite{ibge} provides essential data for this study, including mobility and population information. The mobility data, derived from IBGE's survey of intercity connections, allows the construction of a mobility network representing flows between cities. This network is composed of $N = 5,385$ vertices (cities) and $L = 65,639$ edges (routes), where the weight of each edge corresponds to the frequency of vehicles outgoing between two cities in a typical week. These frequencies encompass various modes of transportation, including roads, waterways, and informal transportation. Attributes such as travel time and cost were available, but were not incorporated into the analysis. In addition, population data from the 2022 Census, provided by IBGE, enhances the mobility network by facilitating the standardization of infection and mortality rates at the municipal level per $100,000$ inhabitants.

\subsubsection{Chinese mobility network}
\label{sec:baidu_data}

The Baidu Mobility Data \cite{China_data_lab_2020} contains daily inflows and outflows of people between origins and destinations within 340 Chinese cities. The two adjacency matrices contain the rate of people who commute from $i$ to $j$, having $j$ as a reference (inflows), and the rate from $j$ to $i$ (outflows). In both cases, considering a reference city, its inflows or outflows sum up to one. In addition, each city has at most 100 neighbors. We aggregated the inflows from January to February 2020 and created a static mobility graph that reflects the average values during this period.

\subsection{COVID-19 time series}

For the Brazilian case, we used the publicly available COVID-19 temporal dataset provided by \cite{wcota}. This dataset contains daily records of infections and deaths at the municipal and state levels in Brazil, sourced from the Ministry of Health. The data span from February 2020 to March 2023, totaling 1095 days. It includes geolocation information for cities, facilitating integration with the mobility network to better understand the dynamics of disease spread.

The Chinese time series covers 694 days from January 19, 2020, to December 12, 2022. The COVID-19 time series contains cases for only 205 of the 340 cities of the mobility data discussed in Section \ref{sec:baidu_data}. We then only used those 205 nodes.

\subsection{Experimental setup}

\subsubsection{Data organization and standardization}

The temporal data were standardized using the \textit{z-score}, a scaler that transforms the values of a variable to a scale with a mean of zero and a standard deviation of one:
\begin{equation}
z = \frac{x - \mu}{\sigma},
\label{eq:zscore}
\end{equation}
where $ x$ is the input value, $\mu$ is the mean, and $\sigma$ is the standard deviation. This standardization ensures that the data are on the same scale, which benefits the models' training.

The graph backbone was extracted for Brazil using the Disparity Filter, reducing the initial number of edges above $65,000$ to just over $20,000$. For this, we use $\alpha=0.01$ and ensure that at least the 5 most weighted connections of each node are retained to ensure that no city is excluded from the analysis. In the end, we have snapshots of the data, where each city (there are $5,385$ in total) is represented by a time series spanning $1,095$ days with information on reported COVID-19 cases.

In the case of China, the graph backbone extraction initially comprised $205$ nodes and $19,046$ edges. Using the same method with identical settings, we achieved a substantial decrease in the number of edges, culminating in slightly more than $980$ connections between all $205$ vertices.

\subsubsection{Train and test data split}

The datasets are transformed into snapshots that serve as input to neural network models, considering a specific window size (context) and a forecast prediction. First, the COVID-19 time series of city-level cases is standardized using a z-score and combined with spatial data on city connectivity and population. Then, snapshots are created, with 80\% allocated to training and the remaining 20\% to testing. Given a temporal dataset covering $1,095$ days, the division for Brazil results in $876$ days for training and $219$ days for testing. For China, of a total of $694$ days, $555$ are for training and $139$ for testing. The application of sliding windows provides flexibility in the number of snapshots generated, making it adaptable to various experimental setups.

\subsubsection{Evaluation metrics}

For regression, we use the Root Mean Squared Error (RMSE):
\begin{equation}
\text{RMSE} = \sqrt{\frac{1}{n} \sum_{i=1}^n (y_i - \hat{y}_i)^2},
\end{equation}
where $y_i$ is the real value, $\hat{y}_i$ is the prediction, and $n$ is the number of samples.

Concerning classification, we use the following:
\begin{itemize}
    \item Precision: proportion of true positives among the predicted positives:
    \begin{equation}
    \text{Precision} = \frac{TP}{TP + FP},
    \end{equation}
    where TP is the number of true positives, and FP represents false positives.
    \item Recall: Proportion of true positives among the actual positive cases:
    \begin{equation}
    \text{Recall} = \frac{TP}{TP + FN},
    \end{equation}
    where $FN$ is the number of false negatives.
    \item F1-Score: Harmonic combination of precision and recall:
    \begin{equation}
    F1 = 2 \cdot \frac{\text{Precision} \cdot \text{Recall}}{\text{Precision} + \text{Recall}}.
    \end{equation}
\end{itemize}

\subsubsection{Experiments}

We build on the results of Duarte \textit{et al.} \cite{duarte2023} by using the Brazilian mobility network and COVID-19 time series of accumulated cases and the neural architectures GCLSTM and GCRN. Initially, the parameters are the same as those used by the authors, with a window size of $14$ days and a prediction horizon of $1$ day. Without employing the sliding window strategy, we end up with a sequence of non-overlapping time series segments of size $15$. The following experiments pile up different strategies to improve the results:
\begin{itemize}
    \item We first reproduce the results of Duarte \textit{et al.} \cite{duarte2023} for accumulated cases in Brazil;
    \item Experiment 1: Next, we include 4 additional months of data (December 2022 to March 2023);
    \item Experiment 2: We add the sliding window strategy, so the number of samples increases for training and testing;
    \item Experiment 3: Finally, we extract the backbone from the mobility network to consider only the most significant neighbors of each node, as described in Section \ref{sec:backbone}.
\end{itemize}

The following experiments are conducted on both the Brazilian and Chinese datasets. Using the last strategy, which employs a sliding window and extracts the network backbone, we now vary both the window size and the prediction horizon from $1$ to $14$ days to capture their dependence. Each snapshot combines temporal and spatial (graph) data for a specific interval. Considering all possible combinations of windows and horizons, we end up with $196$ different cases ($14 \times 14$):
\begin{itemize}
    \item Experiment 4: Regression for Brazil;
    \item Experiment 5: Regression for China;
\end{itemize}

In addition to the regression task of predicting future time series values, we also forecast their tendency via a classification task. This approach is motivated by the work of The Harvard Global Health Institute and Harvard's Edmond J. Safra Center for Ethics \cite{keymetrics}, which has participated in a collaborative effort with a network of research and policy organizations to achieve consensus on key metrics to evaluate the status of response to the pandemic and key performance indicators to assess the effectiveness of response tools deployed. They define risk levels based on new daily cases per 100k population, using a seven-day rolling average and a trend direction and rate. After that, they represent the following four risk levels:
\begin{itemize}
    \item Red: More than 25 daily new cases per 100k people;
    \item Orange: Between 10 and 25 daily new cases per 100k people;
    \item Yellow: Between 1 and 10 daily new cases per 100k people;
    \item Green: Less than 1 daily new case per 100k people.
\end{itemize}

An initial attempt to apply the same method with four classes for each time series of Brazil and China at the city level resulted in an imbalance between the classes, prompting us to modify the approach to encompass just two classes, which we called \textit{Stable} and \textit{Alert}, by adapting the work of \cite{keymetrics}.

For each time series, we calculated the classification targets using a custom function based on the previous seven days of data. The key steps in the target computation are as follows:
\begin{itemize}
    \item Sliding window for moving average:  
    We computed a 7-day moving average for all cities for each day in the dataset. This moving average captures the smoothed trend in the data over the past week, helping to reduce noise in the daily variations.
    \item Normalization per 100k Inhabitants:  
    To make the values comparable across cities with different population sizes, the moving average was normalized to the population size of 100,000 inhabitants. This was achieved by dividing the moving average by the normalized population, which is pre-computed for each city.
    \item Trend calculation using Linear Regression:
    To incorporate the recent trend in the data, we performed a linear regression over the past 7 days for each city. The regression estimated the trend rate as the slope of the best-fit line, representing the rate of change in cases over the window.
    \item Combining metrics:  
    The normalized moving average was then multiplied by the computed trend rate to produce a combined metric for each city. This metric captures both the current level of cases and their recent growth rate.
    \item Binary labeling based on threshold:  
    The combined metric was compared against a predefined threshold of 10. If the metric exceeded the threshold, the city was classified as \textit{Alert} (class 1). Otherwise, it was classified as \textit{Stable} (class 0). This binary classification approach simplifies interpretation and focuses on distinguishing stable situations from alert-worthy ones.
\end{itemize}
By applying this methodology across all cities in the dataset, we produced a consistent and interpretable set of binary classification targets for each day in the time series. This approach effectively captures both the absolute number of cases (normalized for the population) and the recent dynamics of case trends, ensuring that the classification aligns with public health priorities.

The number of samples per class is relatively balanced between Stable and Alert: about 47\% of samples have the Alert label and 53\% Stable in Brazil, while in China it is 45\% Alert and 55\% Stable. The two experiments are the following:
\begin{itemize}
    \item Experiment 6: Classification for Brazil;
    \item Experiment 7: Classification for China.
\end{itemize}

In this stage of the study, the same model previously used for regression is now used for classification. The model has been adapted to predict the two classes (stable and alert) over time series. The output of the classification model is the probability of each class, which is calculated using the logarithm of the softmax function (LogSoftmax).

For regression, the errors are at the city level, so that 
\begin{equation}
\text{RMSE}_i = \sqrt{\frac{1}{n} \sum_{k=1}^{m} \sum_{j=1}^{n}({y}_{i,j} - \hat{y}_{i,j,k})^2},
\end{equation}
for city $i$, with $n$ samples of time series and $m$ model runs. As some architectures are trained 4 times, the RMSE reported here is based on the $m=4$ models. The mean, min, max, and standard deviation are derived from the RMSE values for all cities considered (5,385 for Brazil and 205 for China). 

Lastly, Experiments 8 to 11 were designed to evaluate the performance of the GCLSTM, GCRN, and LSTM models for forecasting daily COVID-19 new cases:
\begin{itemize}
    \item Experiment 8: Regression for Brazil (daily cases);
    \item Experiment 9: Regression for China (daily cases).
    \item Experiment 10: Classification for Brazil (daily cases);
    \item Experiment 11: Classification for China (daily cases).
\end{itemize}

In contrast to previous experiments that focused on cumulative cases, the daily variation data exhibit a significantly imbalanced class distribution: approximately 79\% of samples are labeled as Stable and 21\% as Alert for Brazil, and 99\% are labeled as Stable and 1\% as Alert for China. To address the complexities of this task and enhance the models' convergence and generalization capabilities, we employed:

\begin{itemize}
    \item \textbf{AdamW Optimizer:} The AdamW optimizer (\textit{Adam with Decoupled Weight Decay}) was used, configured with a `weight\_decay` of $1 \times 10^{-3}$, which is effective in regularizing model weights and preventing overfitting.
    \item \textbf{Learning Rate Scheduler:} The learning rate was dynamically adjusted by a combined scheduler. The training began with a 5-epoch \textit{Warmup} phase, during which the learning rate was linearly increased from an initial value of $1 \times 10^{-5}$ to a base rate of $1 \times 10^{-2}$. Following the \textit{Warmup}, a \textit{Cosine Decay} scheduler gradually reduced the learning rate to a minimum value of $1 \times 10^{-4}$ by the end of the total 100 epochs.
    \item \textbf{Early Stopping:} To mitigate overfitting and optimize training time, an early stopping policy was employed. Training was halted if the validation loss did not show a minimum improvement (`min\_delta`) of $0.001$ over 10 consecutive epochs.
\end{itemize}

The models were evaluated using Critical Difference Diagrams based on the Friedman and Nemenyi statistical tests \cite{Rainio_et_al_2024_ml_hypothesis_tests}, which compared the average ranks of the models across different configurations of input window size and prediction horizon.

\begin{figure}[htbp!]
    \centering
    \includegraphics[width=0.9\textwidth]{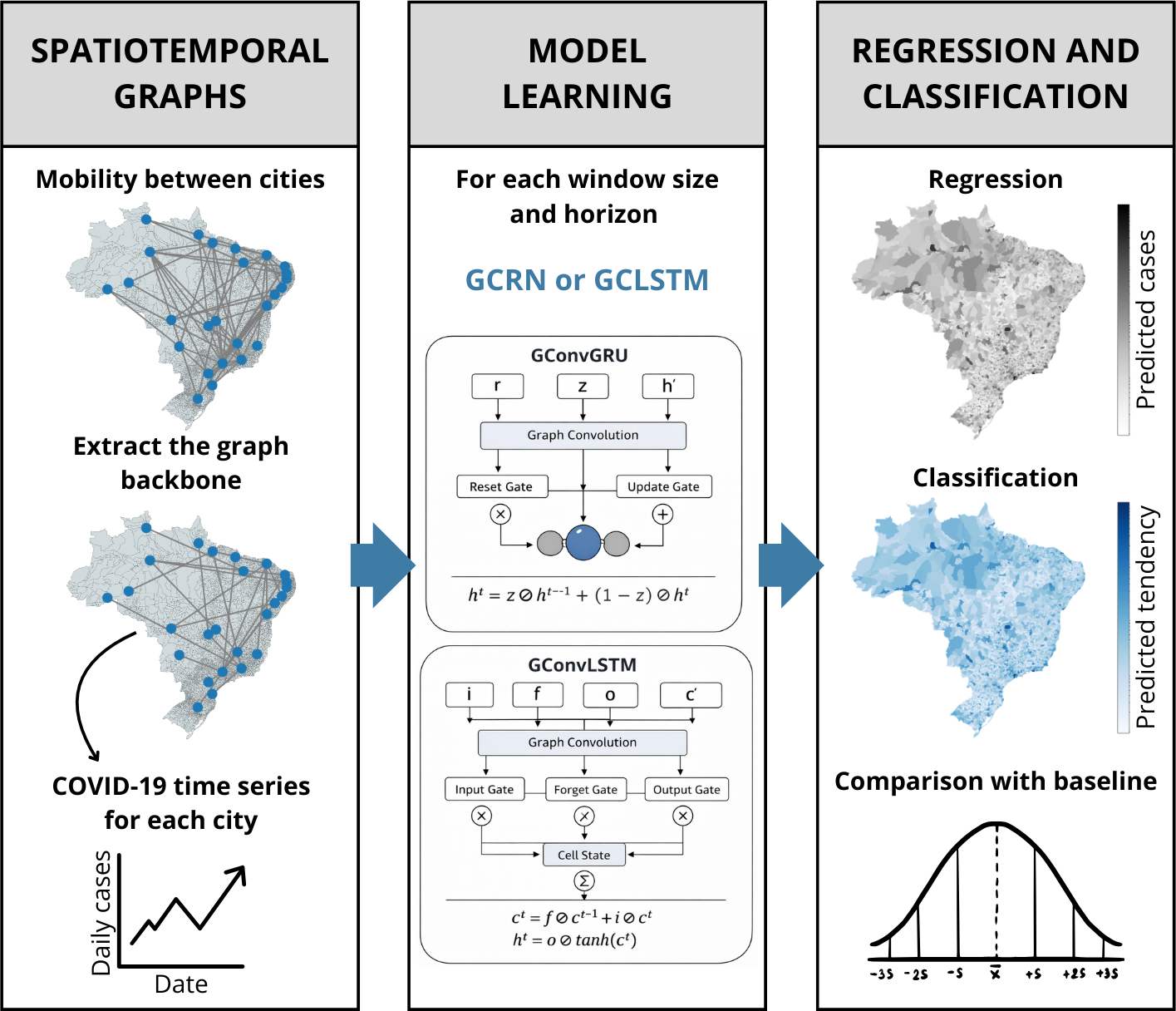}
    \caption{End-to-end pipeline for spatiotemporal COVID-19 modeling: from graph construction and preprocessing to model training and benchmarking against an LSTM baseline.}
    \label{fig:workflow}
\end{figure}

Figure \ref{fig:workflow} summarizes the end-to-end workflow for training and evaluating models on both Brazilian and Chinese datasets. Spatiotemporal graphs are constructed from mobility data and COVID-19 time series, followed by backbone extraction and an 80\%/20\% train-test split. The training data is standardized, with the same transformation applied to the test set. Samples are then generated from a sliding window of fixed size and prediction horizon; in the classification setting, corresponding labels are derived from these windows. Models are trained using either GCRN or GCLSTM, targeting regression or classification tasks. Performance is finally benchmarked against an LSTM baseline, evaluated via RMSE for regression and F1-score, precision, and recall for classification. Table \ref{tab:summary_experiments} presents a summary of the experiments.

\begin{table*}[ht]
\centering
\caption{Summary of experiments. Each experiment builds upon the previous setup, progressively increasing complexity.}
\label{tab:summary_experiments}
\resizebox{\textwidth}{!}{%
\begin{tabular}{c p{3cm} c c c c c c c}
\toprule
\textbf{Experiment} & \textbf{Description} & \textbf{Dataset} & \textbf{Task} & \textbf{Cases} & \textbf{Window} & \textbf{Horizon} & \textbf{Sliding} & \textbf{Backbone} \\
 & & & & & \textbf{Size (days)} & \textbf{(days)} & \textbf{Window} & \\
\midrule
Reproduction       & Reproduces \cite{duarte2023}: accum. cases, fixed window & Brazil & Regression & Accum. & 14 & 1  & \xmark & \xmark \\ \midrule
1                  & Extends data by 4 additional months                          & Brazil & Regression & Accum. & 14 & 1  & \xmark & \xmark \\ \midrule
2                  & Adds sliding window strategy                                 & Brazil & Regression & Accum. & 14 & 1  & \cmark & \xmark \\ \midrule
3                  & Adds backbone extraction                                     & Brazil & Regression & Accum. & 14 & 1  & \cmark & \cmark \\ \midrule
4                  & Multiple window sizes and horizons ($14 \times 14$)          & Brazil & Regression & Accum. & multiple & multiple & \cmark & \cmark \\ \midrule
5                  & Same setup as Experiment 4                                   & China  & Regression & Accum. & multiple & multiple & \cmark & \cmark \\ \midrule
6                  & Classification task                                          & Brazil & Classification & Accum. & multiple & multiple & \cmark & \cmark \\ \midrule
7                  & Classification task                                          & China  & Classification & Accum. & multiple & multiple & \cmark & \cmark \\ \midrule
8                  & Switches from accumulated to daily cases                     & Brazil & Regression & Daily      & multiple & multiple & \cmark & \cmark \\ \midrule
9                  & Same setup as Experiment 8                                   & China  & Regression & Daily      & multiple & multiple & \cmark & \cmark \\ \midrule
10                 & Classification task                                          & Brazil & Classification & Daily      & multiple & multiple & \cmark & \cmark \\ \midrule
11                 & Classification task                                          & China  & Classification & Daily      & multiple & multiple & \cmark & \cmark \\
\bottomrule
\end{tabular}%
}
\end{table*}

\section{Results and discussion}
\label{sec:results}

\subsection{Building on previous results}

Table \ref{tab:model_performance} presents the results for the first three experiments, based on Duarte et al. \cite{duarte2023}. The RMSE is computed for each timestamp in the test set, where each timestamp corresponds to predictions for a set of many cities. This strategy allows us to evaluate the model's error across all cities at each time stamp, capturing its behavior over time. From the resulting RMSE values, we compute the mean, standard deviation, minimum, and maximum to summarize the model's performance over the entire test period. First, one starts without the sliding window technique to have a setup comparable to the work of \cite{duarte2023}, but with 4 more months of data. The results demonstrate a significant reduction in RMSE, with GCLSTM errors dropping from $3535.38$ to $2268.27$. Concerning GCRN, errors dropped from $2990.40$ to $2102.27$. When one adds the sliding window strategy, the amount of data increases, and errors decrease even more, reaching $1150.35$ for GCLSTM and $1315.47$ for GCRN. After backbone extraction, the GCLSTM model achieves an RMSE of $682.14$ ($80.7$\% reduction), and the GCRN model achieves an RMSE of $560.92$ ($81.2$\% reduction). 
\begin{table}[htbp!]
\centering
\caption{Results for GCLSTM and GCRN models with different configurations, using a window size of $14$ and prediction horizon of $1$.}
\label{tab:model_performance}
\resizebox{\textwidth}{!}{
\begin{tabular}{l|l|cccc}
\toprule
\textbf{Model} & \textbf{Scenario} & \textbf{Mean RMSE} & \textbf{Std} & \textbf{Min} & \textbf{Max} \\
\midrule
\multirow{4}{*}{GCLSTM} & Duarte \textit{et al.} \cite{duarte2023} & 3535.38 & 1179.61 & 1337.55 & 5327.53 \\
                        & Exp. 1: more data                     & 2268.27 & 1189.09 & 1050.93 & 4564.23 \\
                        & Exp. 2: sliding window                & 1150.35 & 924.28  & 443.17  & 4549.68 \\
                        & Exp. 3: sliding window; backbone      & 682.14  & 950.88  & 249.75  & 4059.08 \\
\midrule
\multirow{4}{*}{GCRN}   & Duarte \textit{et al.} \cite{duarte2023} & 2990.40 & 1000.01 & 1178.76 & 4598.27 \\
                        & Exp. 1: more data                     & 2102.27 & 1336.78 & 804.79  & 4743.18 \\
                        & Exp. 2: sliding window                & 1315.47 & 852.58  & 709.11  & 4560.40 \\
                        & Exp. 3: sliding window; backbone      & 560.92  & 875.39  & 150.23  & 4173.21\\
                        \bottomrule
\end{tabular}}
\end{table}

There is progressive improvement in results as new techniques are applied, highlighting the effectiveness of sliding-window and backbone-extraction strategies.

\subsection{Regression performance for accumulated cases}

\begin{figure}[htbp!]
    \centering
    \includegraphics[width=0.95\textwidth]{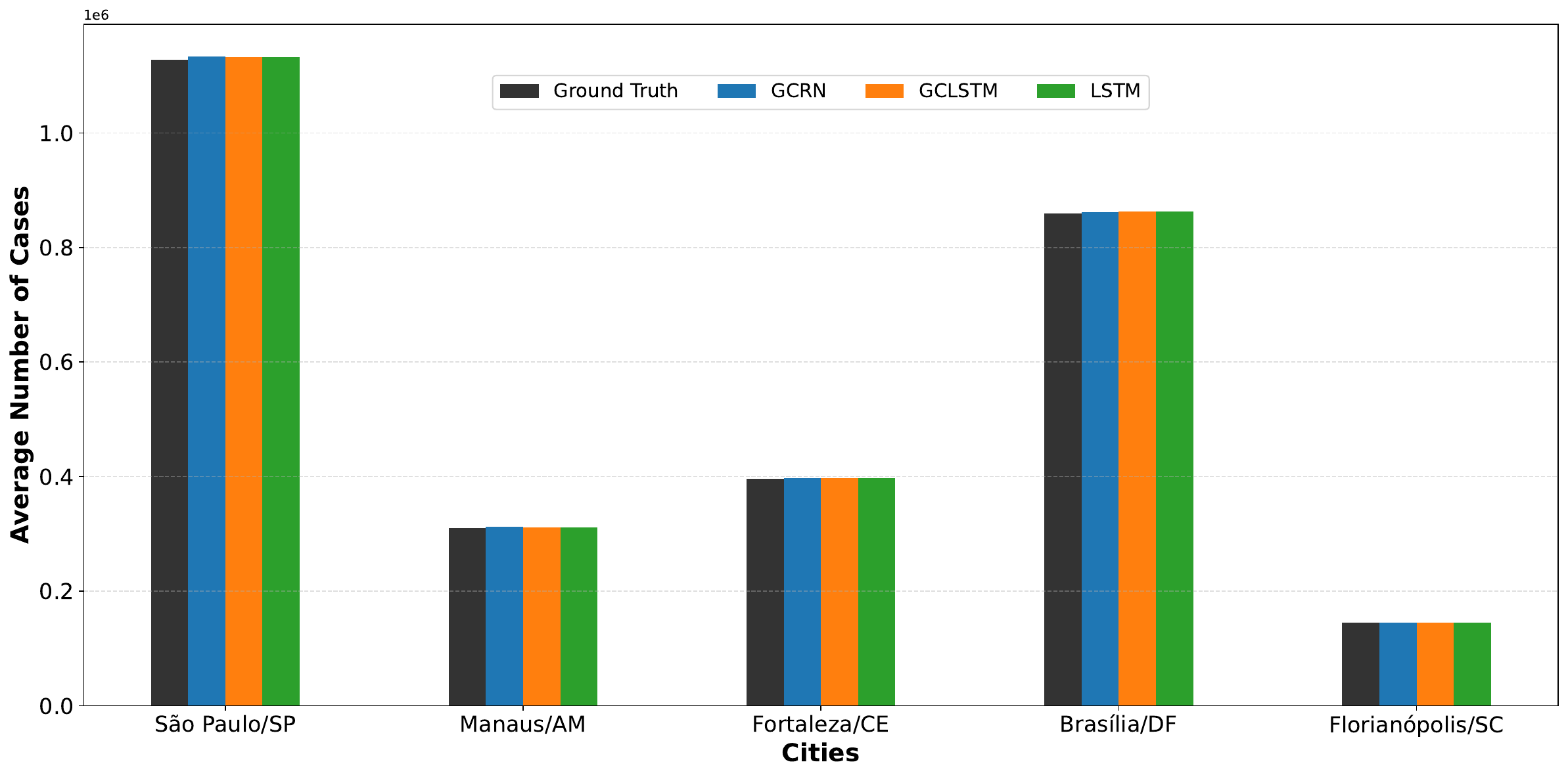}
    \caption{Comparison of the overall average predicted cases versus the ground truth for selected cities in Brazil. The values are averaged across all combinations of window size and prediction horizon (1-14).}
    \label{fig:comparison_br}
\end{figure}

\begin{figure}[htbp!]
    \centering
    \includegraphics[width=0.95\textwidth]{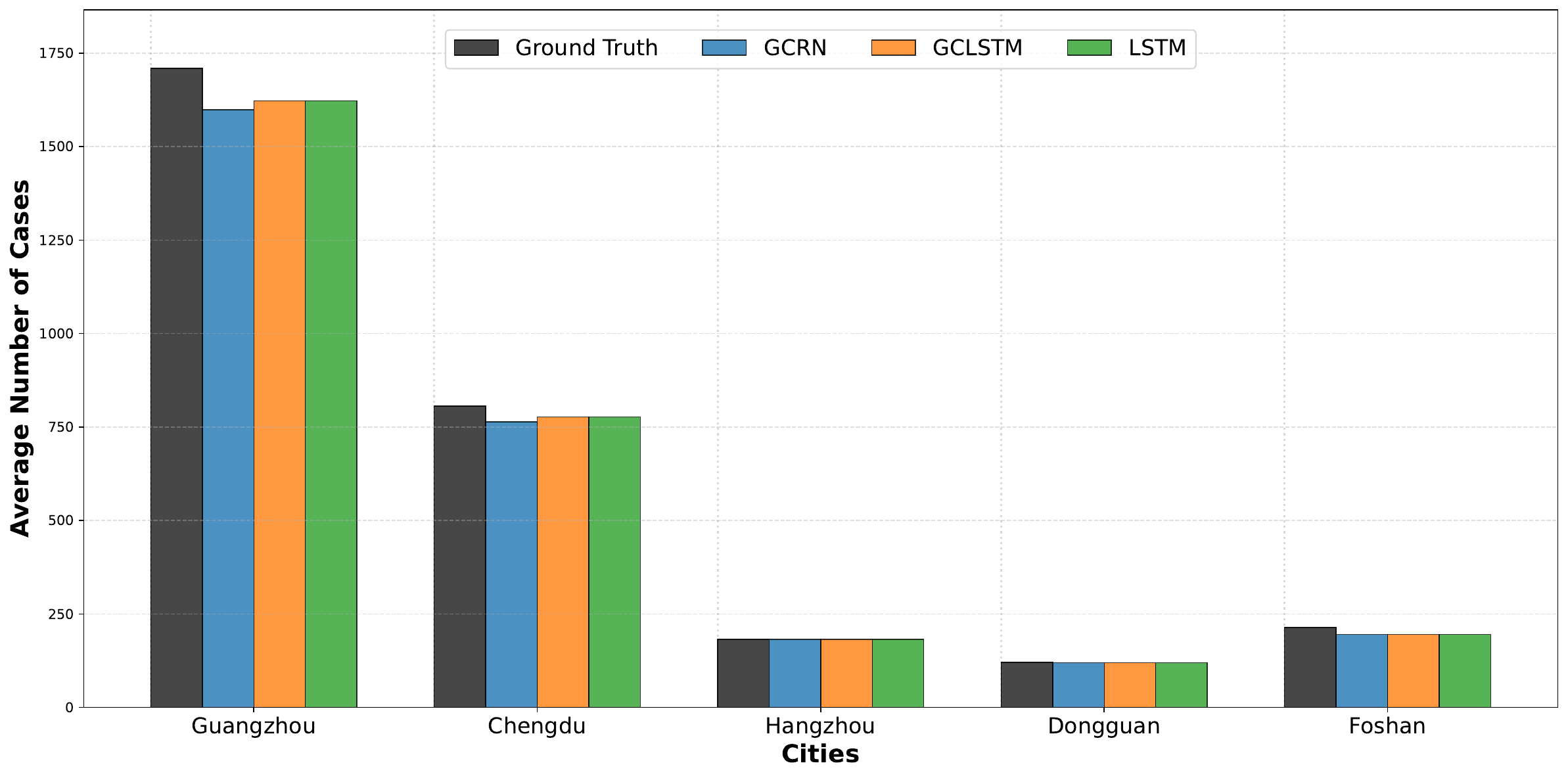}
    \caption{Comparison of the overall average predicted cases versus the ground truth for selected cities in China. The values are averaged across all combinations of window size and prediction horizon (1-14).}
    \label{fig:comparison_ch}
\end{figure}

Figures \ref{fig:comparison_br} and \ref{fig:comparison_ch} present a direct comparison between the overall average predictions of each model and the ground truth values for selected cities in Brazil and China, respectively. These figures are provided for the reader's reference and illustrate the models' average performance across all 196 tested configurations of window size and prediction horizon.

\begin{figure}[htpb!]
    \centering
    \includegraphics[width=\textwidth]{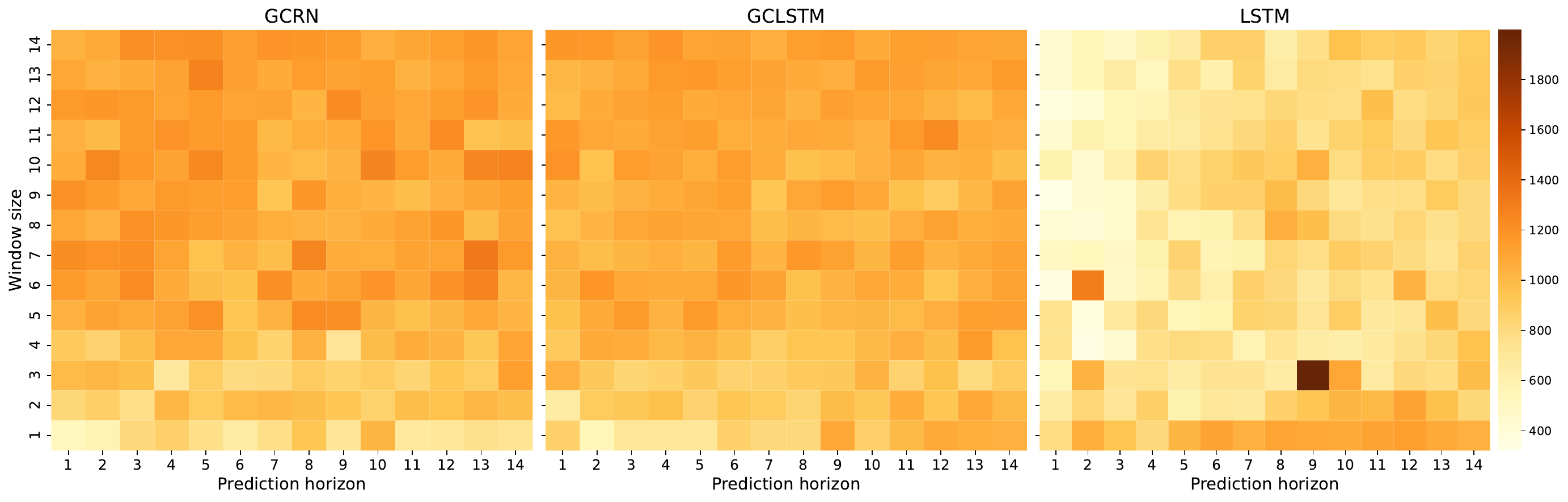}
    \caption{Experiment 4 (Brazil) - average RMSE for regression (GCRN, GCLSTM, LSTM; Window/Horizon 1-14).}
    \label{fig:heatmap_rmse_br}
\end{figure}

Figure \ref{fig:heatmap_rmse_br} presents the comparative results of Experiment 4 between the GCRN, GCLSTM, and LSTM models in regression tasks, considering different window sizes ($1-14$) and prediction horizons ($1-14$) for Brazil. 

\begin{table}[htpb!]
\small
\centering
\caption{Experiment 4 - Performance Comparison between Models in Brazil}
\label{tab:regression_comparation_models_br}
\begin{tabular}{l|c|c|c}
\toprule
Statistic & GCRN & GCLSTM & LSTM\\
\midrule
Maximum RMSE & 1319.05 & \textbf{1228.81} & 1997.74\\
Minimum RMSE & 520.92 & 548.02 & \textbf{320.93}\\
Mean RMSE & 1048.06 & 1030.66 & \textbf{768.95}\\
Standard Deviation RMSE & 140.95 & \textbf{106.02} & 202.16\\
1st Quartile RMSE & 984.12 & 979.20 & \textbf{647.52}\\
Median RMSE & 1072.72 & 1053.12 & \textbf{780.78}\\
3rd Quartile RMSE & 1147.40 & 1101.07 & \textbf{869.26}\\
\bottomrule
\end{tabular}
\end{table}

In the regression task for the Brazilian dataset, the LSTM model demonstrated a clear superiority in overall performance. As shown in Table \ref{tab:regression_comparation_models_br}, the LSTM achieved a significantly lower Mean RMSE (768.95) compared to both GCRN (1048.06) and GCLSTM (1030.66). This performance gap is also visually apparent in Figure \ref{fig:heatmap_rmse_br}, where the heatmap for LSTM is predominantly lighter, indicating lower error across most configurations.

A deeper analysis of the LSTM's performance reveals that it not only achieved the lowest mean error but also the overall minimum RMSE (320.93), showcasing its potential for high accuracy. However, its predictions exhibited greater variability, evidenced by the largest standard deviation (202.16) and the highest maximum error among the three models. This suggests that while the standard LSTM is highly accurate on average, it may be susceptible to occasional larger errors compared to the more constrained GNN models in this regression context.

In contrast, the GNN-based models, GCRN and GCLSTM, showed remarkably similar performance. The slight difference in their mean RMSE values ($\Delta = 17.40$) indicates comparable predictive capabilities on average. The quartile analysis confirms this, revealing a very similar error distribution. However, the GCLSTM proved slightly more consistent, with a lower standard deviation (106.02) and a narrower error range (680.79) compared to the GCRN's standard deviation (140.95) and error range (798.13). This suggests that while both GNN models perform comparably, the GCLSTM offers a marginally more stable performance.

A common trend across all three models is a decline in performance as the prediction horizon increases, highlighting the inherent challenge of forecasting farther into the future.

\begin{figure}[htbp!]
    \centering
    \includegraphics[width=\textwidth]{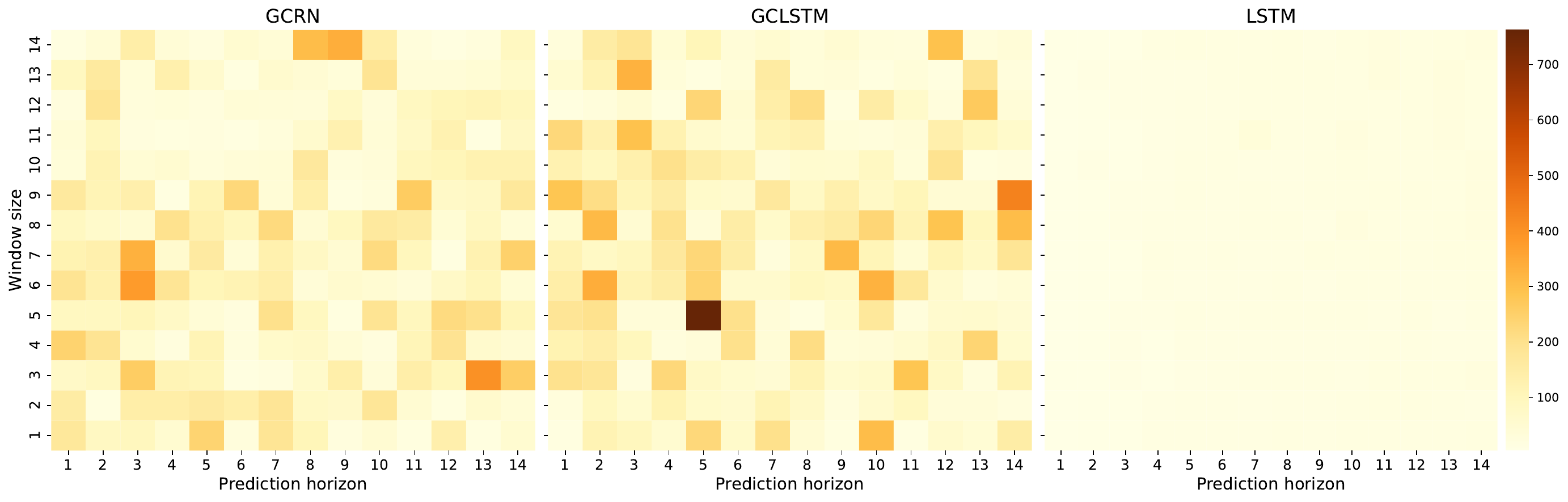}
    \caption{Experiment 5 - Average RMSE Heatmaps for Regression Task, China.}
    \label{fig:heatmap_rmse_ch}
\end{figure}

Figure \ref{fig:heatmap_rmse_ch} shows the results for the Chinese dataset. For this case, the performance disparity was even more pronounced, with the LSTM model again significantly outperforming both GNN-based models. The LSTM heatmap indicates consistently low errors across nearly all configurations. In contrast, the GCRN and GCLSTM models display more heterogeneous heatmaps with noticeable ``hotspots'' of high error, such as the outlier observed for GCLSTM at a window size and prediction horizon of $5$. This outcome suggests that for the Chinese time series, which may possess different underlying patterns, the spatial information captured by the GNNs did not translate into a predictive advantage for the regression task. Instead, the standard LSTM's strong sequential modeling capability has proven more effective.

\begin{table}[htbp!]
\small
\centering
\caption{Performance Comparison between Models in China}
\label{tab:regression_comparation_models_china}
\begin{tabular}{l|c|c|c}
\toprule
Statistic & GCRN & GCLSTM & LSTM\\
\midrule
Maximum RMSE & 401.35 & 762.37 & 31.56\\
Minimum RMSE & 16.22 & 16.41 & 4.01\\
Mean RMSE & 96.65 & 109.09 & 15.13\\
Standard Deviation RMSE & 73.81 & 95.55 & 4.82\\
1st Quartile RMSE & 40.02 & 38.65 & 11.97\\
Median RMSE & 80.52 & 77.65 & 15.90\\
3rd Quartile RMSE & 133.75 & 148.93 & 18.07\\
\bottomrule
\end{tabular}
\end{table}

In Table \ref{tab:regression_comparation_models_china}, the GCRN and GCLSTM models show similar statistical characteristics for the Chinese results. The most noticeable difference lies in the maximum RMSE values, where the GCLSTM model shows a significantly higher maximum error ($762.37$) than the GCRN model ($401.35$). This suggests that while both models perform similarly under most conditions, the GCLSTM has occasional predictions with substantially higher errors.

The standard deviation further highlights the models' different error distributions. The GCRN's lower standard deviation of $73.81$ indicates more consistent predictions, while the GCLSTM's higher standard deviation of $95.55$ suggests more variable performance. The GCRN model showed a more constrained range of RMSE values, spanning from $16.22$ to $401.35$, with a mean of $96.65$ and a standard deviation of $73.81$. In contrast, the GCLSTM model exhibited more significant variability, with RMSE values ranging from $16.41$ to $762.37$, a mean of $109.09$, and a standard deviation of $95.55$. 

\subsection{Classification performance for accumulated cases}

\begin{figure}[htbp!]
    \centering
    \includegraphics[width=\textwidth]{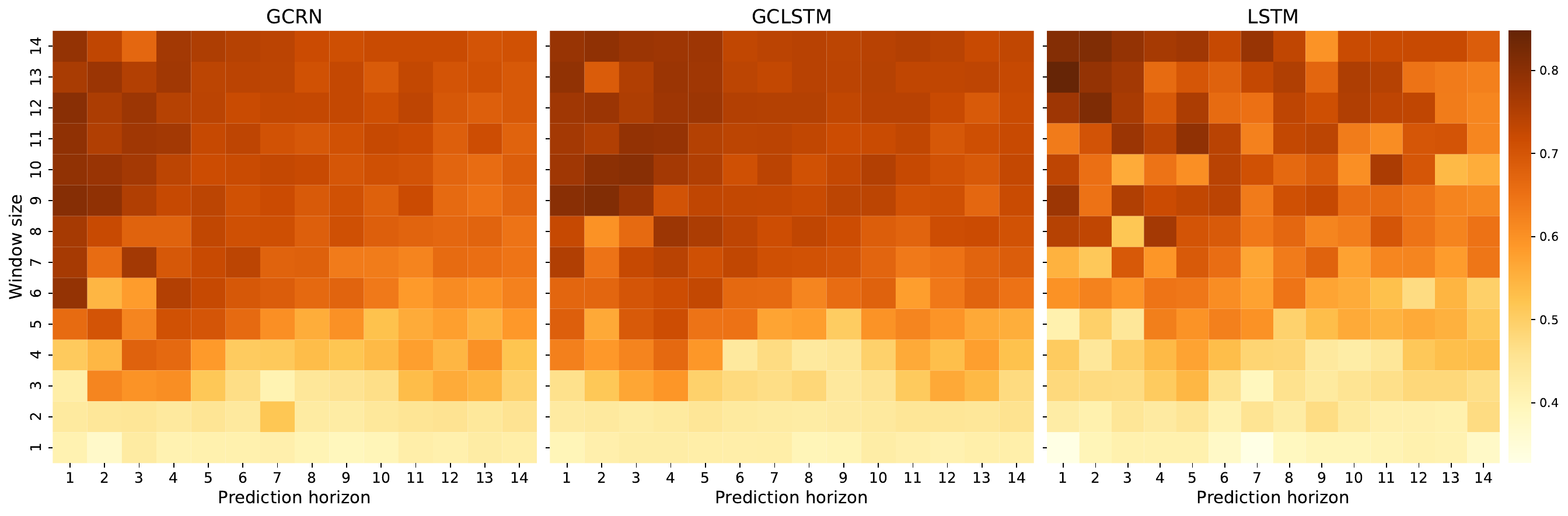}
    \caption{Experiment 6 - Average F1-Score Heatmaps for Classification Task, Brazil.}
    \label{fig:heatmap_f1_score_br}
\end{figure}

\begin{figure}[htbp!]
    \centering
    \includegraphics[width=\textwidth]{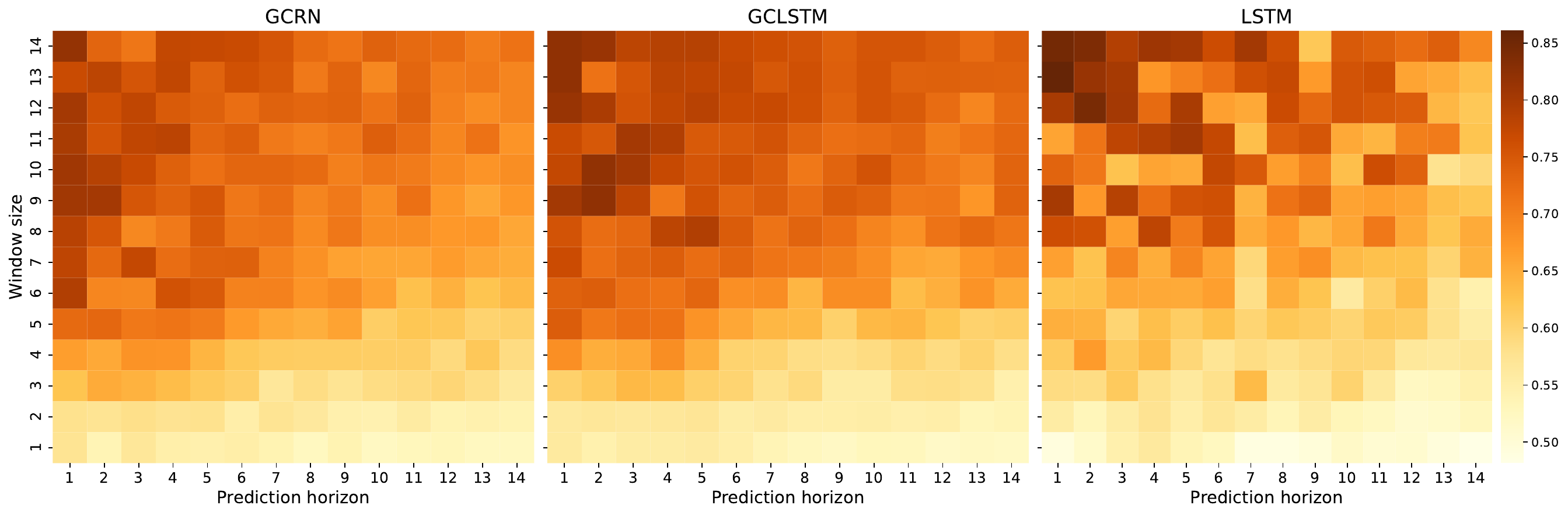}
    \caption{Experiment 6 - Average Precision Heatmaps for Classification Task, Brazil.}
    \label{fig:heatmap_precision_br}
\end{figure}

\begin{figure}[htbp!]
    \centering
    \includegraphics[width=\textwidth]{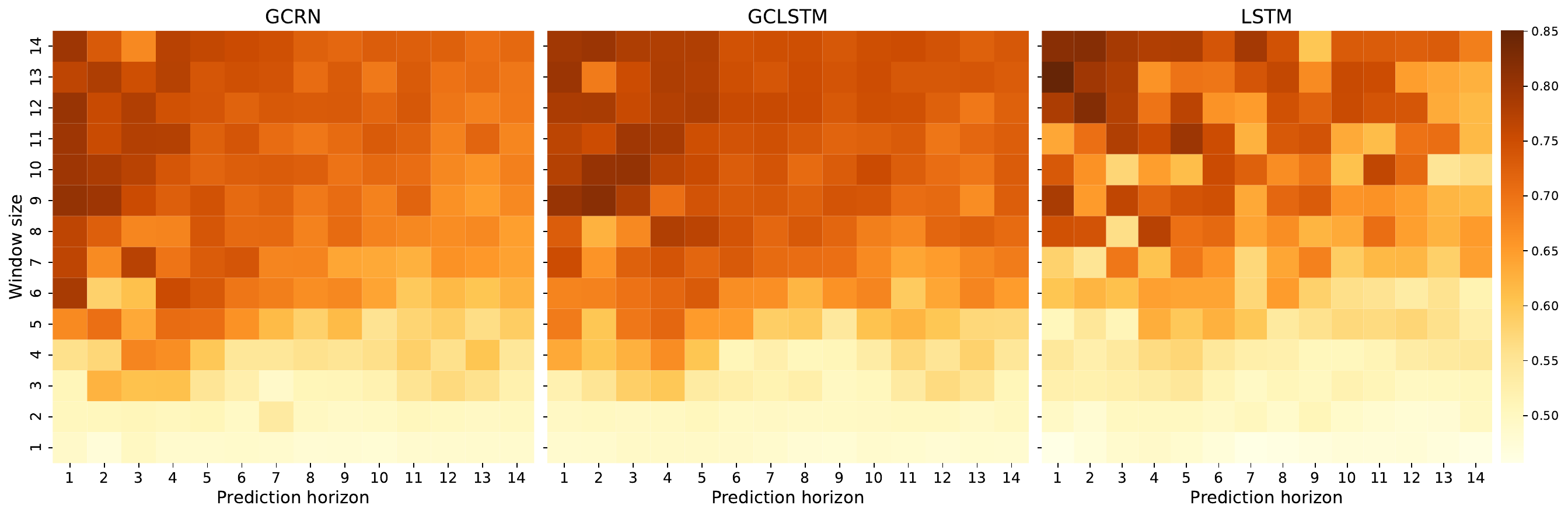}
    \caption{Experiment 6 - Average Recall Heatmaps for Classification Task, Brazil.}
    \label{fig:heatmap_recall_br}
\end{figure}

Regarding Experiment 6, classification with the Brazilian dataset, both models perform better for shorter prediction horizons with larger window sizes. Observing the heatmaps for the F1-Score, Precision, and Recall metrics (Figures \ref{fig:heatmap_f1_score_br}, \ref{fig:heatmap_precision_br}, and \ref{fig:heatmap_recall_br}), this outcome was expected. The GCRN model achieves maximum scores of $86$\% for F1-Score, Precision, and Recall (window size $8$, prediction horizon $1$), while the GCLSTM reaches $83$\% (window size $3$, prediction horizon $1$). For window sizes of $5$ or more, improvements across all metrics are observed, with the GCRN consistently peaking at $8$ with a prediction horizon of $1$. Some key observations are:
\begin{itemize}
    \item GCRN demonstrates more stable performance across different configurations.
    \item GCLSTM exhibits slightly greater volatility in performance.
    \item Both models show a degradation in performance as the prediction horizon increases.
    \item Smaller window sizes (1-4) produce poorer results for both models.
    \item The worst results are close to 50\%.
    \item Window sizes between 6-10 yield the best results for both models.
\end{itemize}

Table \ref{tab:classification_metrics_br} presents a general comparison of classification performance metrics for the GCRN and GCLSTM models, highlighting the best values among the models.
\begin{table}[htbp!]
%\scriptsize
\centering
\caption{Experiment 6 - Classification Metrics (Brazil)}
\label{tab:classification_metrics_br}
\begin{tabular}{cl|ccc}
\toprule
\multicolumn{2}{c|}{Metric} & GCRN & GCLSTM & LSTM\\
\midrule
\multirow{7}{*}{\textbf{F1-Score}}&Maximum & 0.86 & 0.83 & 0.85\\
&Minimum & 0.35 & 0.35 & 0.33\\
&Mean & \textbf{0.64} & \textbf{0.64} & 0.60\\
&Std. Dev. & 0.12 & 0.13 & 0.12\\
&1st Quartile & 0.55 & 0.53 & 0.49\\
&Median & 0.68 & 0.69 & 0.62\\
&3rd Quartile & 0.73 & 0.74 & 0.70\\
\midrule
\multirow{7}{*}{\textbf{Precision}}&Maximum & 0.86 & 0.84 & 0.86 \\
&Minimum & 0.48 & 0.47 & 0.48 \\
&Mean & 0.68 & \textbf{0.69} & 0.65 \\
&Std. Dev. & 0.08 & 0.09 & 0.09 \\
&1st Quartile & 0.61 & 0.61 & 0.58 \\
&Median & 0.69 & 0.71 & 0.64 \\
&3rd Quartile & 0.74 & 0.76 & 0.72 \\
\midrule
\multirow{7}{*}{\textbf{Recall}}&Maximum & 0.86 & 0.83 & 0.85 \\
&Minimum & 0.46 & 0.45 & 0.46 \\
&Mean & 0.65 & \textbf{0.66} & 0.62 \\
&Std. Dev. & 0.10 & 0.11 & 0.10 \\
&1st Quartile & 0.57 & 0.56 & 0.53 \\
&Median & 0.68 & 0.69 & 0.62 \\
&3rd Quartile & 0.73 & 0.75 & 0.71\\
\bottomrule
\end{tabular}
\end{table}

Starting with the F1-Score, the GCRN model achieves a maximum value ($0.86$) compared to the GCLSTM ($0.83$). This pattern is consistent across Precision and Recall analyses. The same holds true for the minimum values, means, and standard deviations in the three metrics, indicating a balance of performance between the models. Quartile statistics reveal a very similar distribution between the two models, with a median of $0.68$ for GCRN and $0.69$ for GCLSTM.

The GCRN and GCLSTM models showed very similar performance in all metrics analyzed. However, GCLSTM demonstrates slightly better average performance (mean and median) in precision and recall. However, the GCRN exhibits greater performance stability, with a lower standard deviation across all metrics.

\begin{figure}[htbp!]
    \centering
    \includegraphics[width=\textwidth]{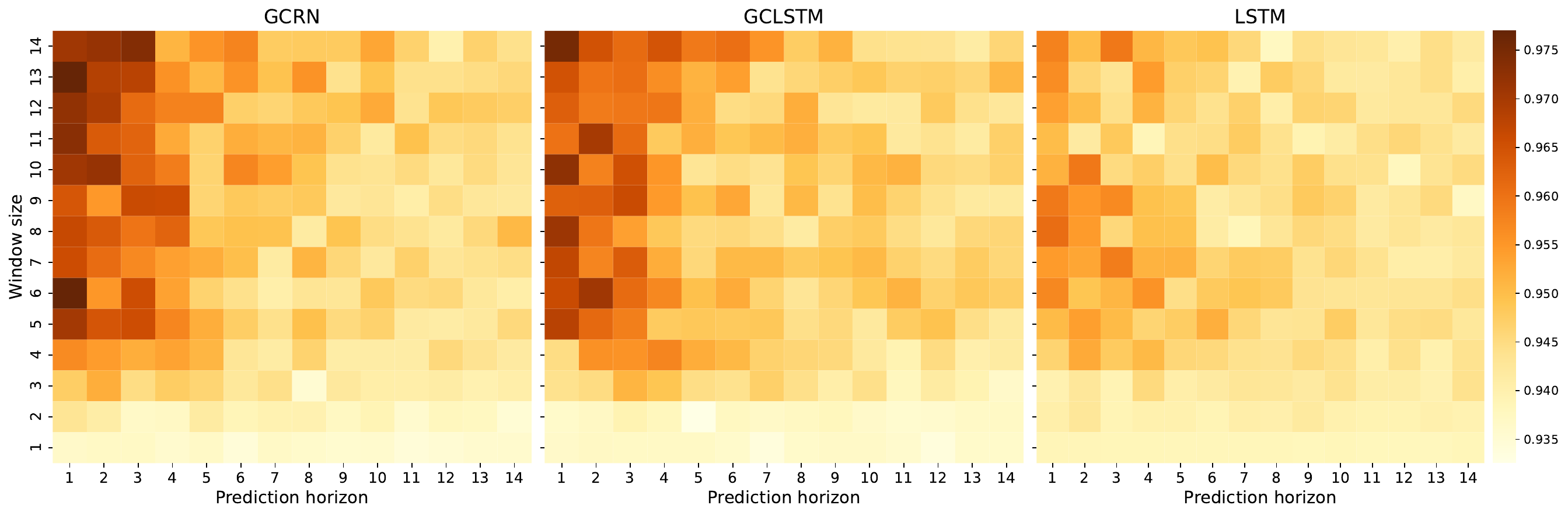}
    \caption{Experiment 7 - Average F1-Score Heatmaps for Classification Task, China.}
    \label{fig:heatmap_f1_score_ch}
\end{figure}

\begin{figure}[htbp!]
    \centering
    \includegraphics[width=\textwidth]{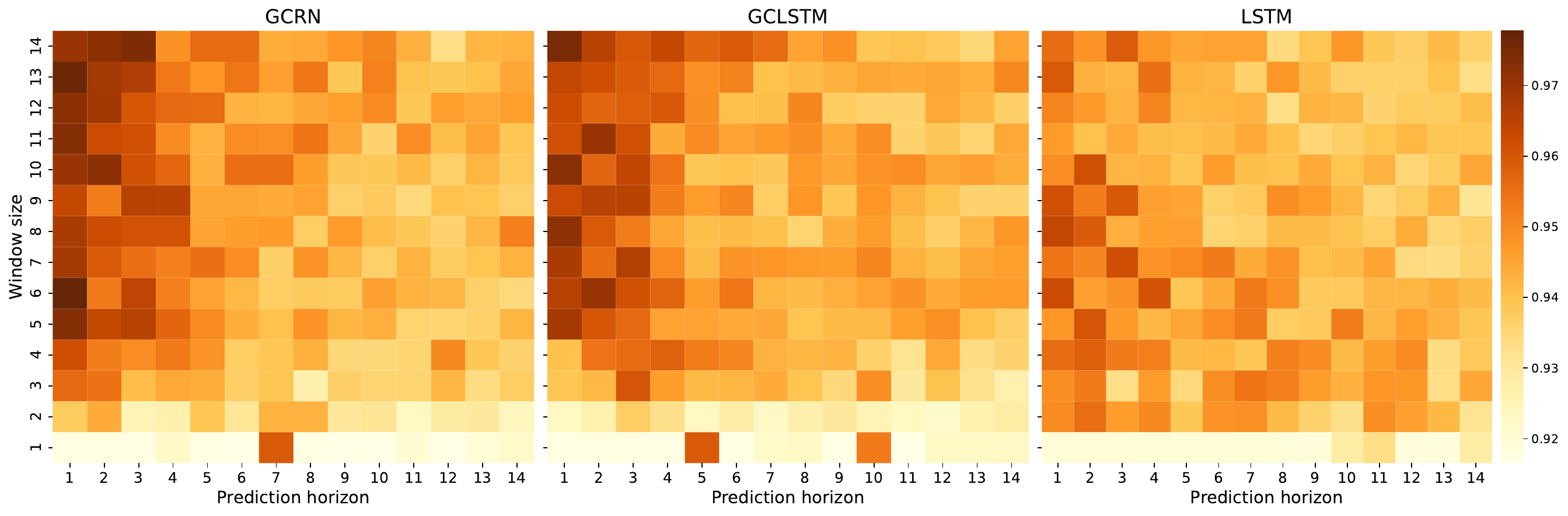}
    \caption{Experiment 7 - Average Precision Heatmaps for Classification Task, China.}
    \label{fig:heatmap_precision_ch}
\end{figure}

\begin{figure}[htbp!]
    \centering
    \includegraphics[width=\textwidth]{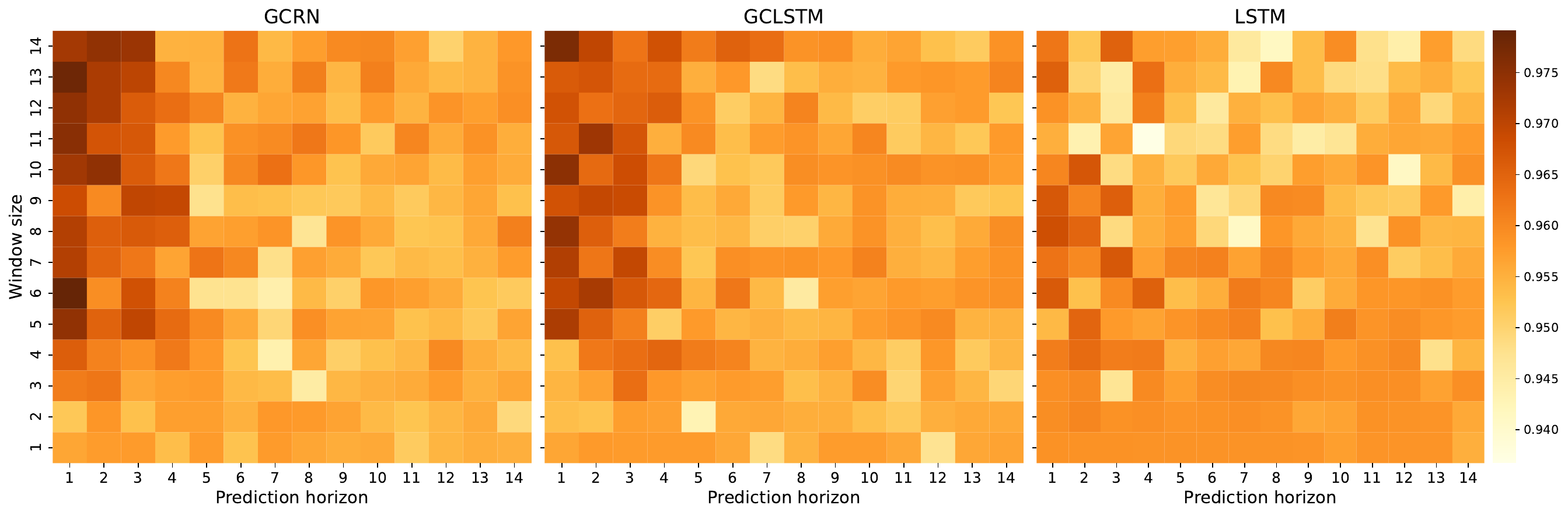}
    \caption{Experiment 7 - Average Recall Heatmaps for Classification Task, China.}
    \label{fig:heatmap_recall_ch}
\end{figure}

Concerning Experiment 7, the performance analysis of the GCRN and GCLSTM models for the China dataset reveals in the heatmaps for F1-score (Figure \ref{fig:heatmap_f1_score_ch}), precision (Figure \ref{fig:heatmap_precision_ch}), and recall (Figure \ref{fig:heatmap_recall_ch}) performance values ranging between $0.94$ and $0.98$ for both models.

Table \ref{tab:classification_metrics_cn} presents a general comparison of classification performance metrics for the GCRN and GCLSTM models, highlighting the best values among the models.
\begin{table}[htbp!]
%\scriptsize
\centering
\caption{Experiment 7 - Classification Metrics (China)}
\label{tab:classification_metrics_cn}
\begin{tabular}{cl|ccc}
\toprule
\multicolumn{2}{c|}{Metric} & GCRN & GCLSTM & LSTM \\
\midrule
\multirow{7}{*}{\textbf{F1-Score}}&Maximum & 0.98 & 0.98 & 0.96 \\
&Minimum & 0.93 & 0.93 & 0.94 \\
&Mean & 0.95 & 0.95 & 0.94 \\
&Std. Dev. & 0.01 & 0.01 & 0.01 \\
&1st Quartile & 0.94 & 0.94 & 0.94 \\
&Median & 0.95 & 0.95 & 0.94 \\
&3rd Quartile & 0.95 & 0.95 & 0.95 \\
\midrule
\multirow{7}{*}{\textbf{Precision}}&Maximum & \textbf{0.98} & 0.97 & 0.96 \\
&Minimum & 0.92 & 0.92 & 0.92 \\
&Mean & 0.94 & 0.94 & 0.94 \\
&Std. Dev. & 0.01 & 0.01 & 0.01 \\
&1st Quartile & 0.94 & 0.94 & 0.94 \\
&Median & 0.94 & 0.94 & 0.94 \\
&3rd Quartile & 0.95 & 0.95 & 0.95 \\
\midrule
\multirow{7}{*}{\textbf{Recall}}&Maximum & 0.98 & 0.98 & 0.97 \\
&Minimum & 0.94 & 0.94 & 0.94 \\
&Mean & 0.96 & 0.96 & 0.96 \\
&Std. Dev. & 0.01 & 0.01 & 0.01 \\
&1st Quartile & 0.95 & 0.95 & 0.95 \\
&Median & 0.96 & 0.96 & 0.96 \\
&3rd Quartile & 0.96 & 0.96 & 0.96\\
\bottomrule
\end{tabular}
\end{table}

The average F1-score of the GCLSTM remains consistently at $0.96$, with a standard deviation of only $0.01$, while the GCRN shows a variation of $0.02$. In precision metrics, a gradual reduction of approximately $3$\% is observed in longer prediction horizons, with the GCLSTM decreasing from $0.97$ to $0.94$ and the GCRN from $0.96$ to $0.92$. This decline is particularly pronounced after the tenth prediction horizon, highlighting the increasing challenges in long-term forecasting.

The comparative analysis of the two models suggests that both are effective for the classification task in the China dataset. The GCLSTM stands out for its more uniform performance, with a coefficient of variation $33$\% lower than that of the GCRN. The results demonstrate statistical robustness, with $92$\% of predictions remaining above $0.94$, indicating significant adaptability to the specific classification context of Chinese data.

\subsection{Daily Cases in Brazil}

So far, results based on accumulated cases suggest that LSTM is superior to GNN-based models. However, Experiment 8, which focuses on the regression analysis of daily COVID-19 cases in Brazil, reveals a significant shift in performance. Daily case variations yield a much less stable dataset than the accumulated series, challenging models in different ways.

In this specific context for Brazilian data, the GCRN model demonstrated superiority in regression tasks, achieving the best performance across the majority of the evaluated scenarios. Table~\ref{tab:brasil_results_exp8_reg_stats} presents the descriptive statistics for the RMSE metric across all configurations. The GCRN model achieves a lower Mean RMSE (587.16) than the LSTM (595.23) and lower Median RMSE (589.22) than the baseline (597.11), demonstrating the consistency and predictive accuracy of the graph-based approach in handling the daily variations.
\begin{table}[htbp!]
\small
\centering
\caption{Experiment 8 - Performance Comparison between Models in Brazil (Regression - RMSE)}
\label{tab:brasil_results_exp8_reg_stats}
\begin{tabular}{l|c|c|c}
\toprule
Statistic & GCRN & GCLSTM & LSTM\\
\midrule
Maximum RMSE & 593.38 & 593.73 & 611.06\\
Minimum RMSE & 566.21 & 569.18 & 560.78\\
Mean RMSE & \textbf{587.16} & 588.16 & 595.23\\
Standard Deviation RMSE & 6.21 & 5.26 & 7.87\\
1st Quartile RMSE & 586.03 & 587.24 & 590.08\\
Median RMSE & 589.22 & 589.86 & 597.11\\
3rd Quartile RMSE & 590.92 & 591.30 & 600.91\\
\bottomrule
\end{tabular}
\end{table}

\begin{figure}[htbp!]
    \centering
    \includegraphics[width=\textwidth]{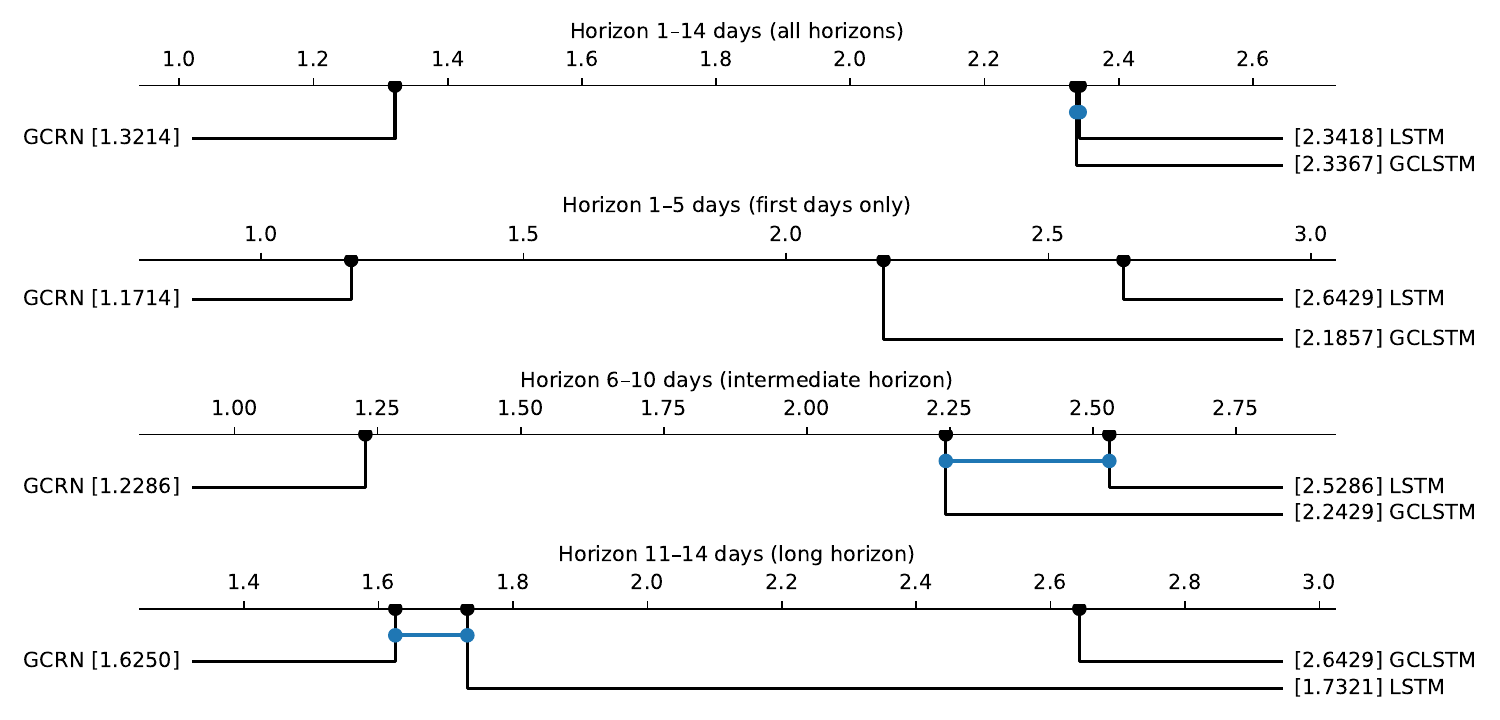}
    \caption{Experiment 8 - Results for Brazil: RMSE regression metric across all horizons. Horizontal blue lines connecting different models indicate that their performances are not statistically different.}
    \label{fig:br_rmse_all}
\end{figure}

The Critical Difference Diagrams for regression, presented in Figure~\ref{fig:br_rmse_all}, corroborate these findings across all evaluated horizons. The diagrams illustrate the average rank for each model, with lower ranks indicating better performance. GCRN consistently appears as the top-ranked model (lowest rank values), reinforcing its robustness in minimizing prediction errors. 
%
% Figura de RMSE

Experiment 9 focused on the classification performance for daily cases in Brazil. In this task (predicting the rise or fall of cases), a distinct trade-off was observed between the models. Table~\ref{tab:classification_metrics_br_exp9} highlights the detailed statistics for each classification metric.

The GCRN model demonstrated superior Precision, achieving a Mean of 0.64 and a Median of 0.61, which were significantly higher than those of the LSTM (Mean 0.56, Median 0.56). 
The high Precision is a critical indicator here, suggesting that the graph-based model is highly effective at minimizing false positives, ensuring that alerts are raised with higher confidence. Conversely, the LSTM model dominated in Recall (Mean 0.36 vs 0.24 for GCRN) and consequently in F1-Score (Mean 0.41 vs 0.34 for GCRN). This indicates that the LSTM is more sensitive to the positive class (rising cases), albeit at the cost of significantly lower precision than the GCRN.
\begin{table}[htbp!]
\centering
\caption{Experiment 9 - Classification Metrics Statistics (Brazil)}
\label{tab:classification_metrics_br_exp9}
\begin{tabular}{cl|ccc}
\toprule
\multicolumn{2}{c|}{Metric} & GCRN & GCLSTM & LSTM \\
\midrule
\multirow{7}{*}{\textbf{F1-Score}}&Maximum & 0.63 & 0.63 & \textbf{0.71} \\
&Minimum & 0.07 & 0.07 & 0.07 \\
&Mean & 0.34 & 0.34 & \textbf{0.41} \\
&Std. Dev. & 0.12 & 0.12 & 0.14 \\
&1st Quartile & 0.28 & 0.28 & 0.35 \\
&Median & 0.33 & 0.33 & 0.41 \\
&3rd Quartile & 0.41 & 0.41 & 0.50 \\
\midrule
\multirow{7}{*}{\textbf{Precision}}&Maximum & 0.82 & 0.82 & \textbf{0.86} \\
&Minimum & \textbf{0.55} & 0.55 & 0.33 \\
&Mean & \textbf{0.64} & 0.63 & 0.56 \\
&Std. Dev. & 0.07 & 0.07 & 0.11 \\
&1st Quartile & 0.58 & 0.58 & 0.48 \\
&Median & \textbf{0.61} & 0.61 & 0.56 \\
&3rd Quartile & 0.67 & 0.67 & 0.62 \\
\midrule
\multirow{7}{*}{\textbf{Recall}}&Maximum & 0.51 & 0.52 & \textbf{0.63} \\
&Minimum & 0.04 & 0.04 & 0.04 \\
&Mean & 0.24 & 0.24 & \textbf{0.36} \\
&Std. Dev. & 0.11 & 0.11 & 0.15 \\
&1st Quartile & 0.18 & 0.18 & 0.26 \\
&Median & 0.23 & 0.23 & 0.39 \\
&3rd Quartile & 0.30 & 0.30 & 0.46\\
\bottomrule
\end{tabular}
\end{table}

\begin{figure}[htbp!]
    \centering
    \includegraphics[width=\textwidth]{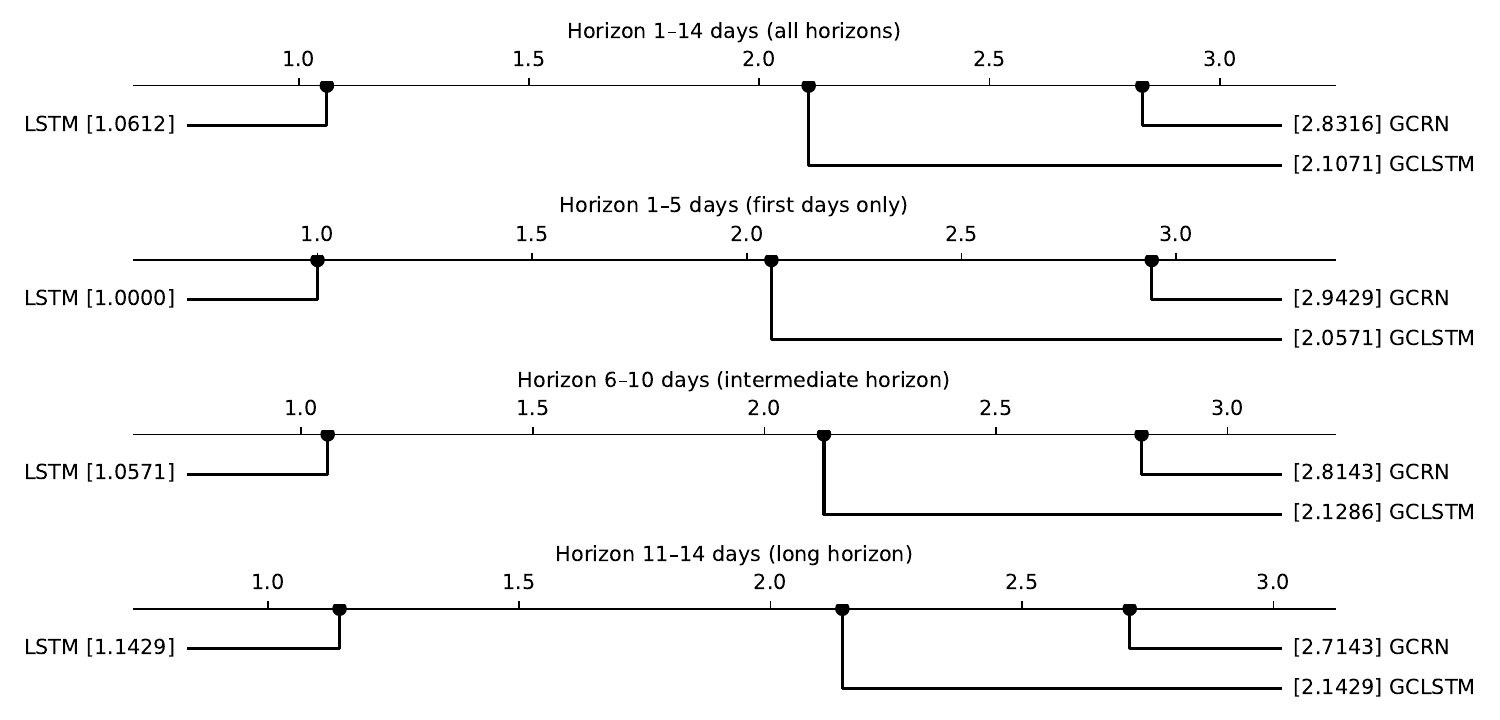}
    \caption{Experiment 9 - Results for Brazil: F1-Score classification metric across all horizons.}
    \label{fig:br_f1}
\end{figure}

\begin{figure}[htbp!]
    \centering
    \includegraphics[width=\textwidth]{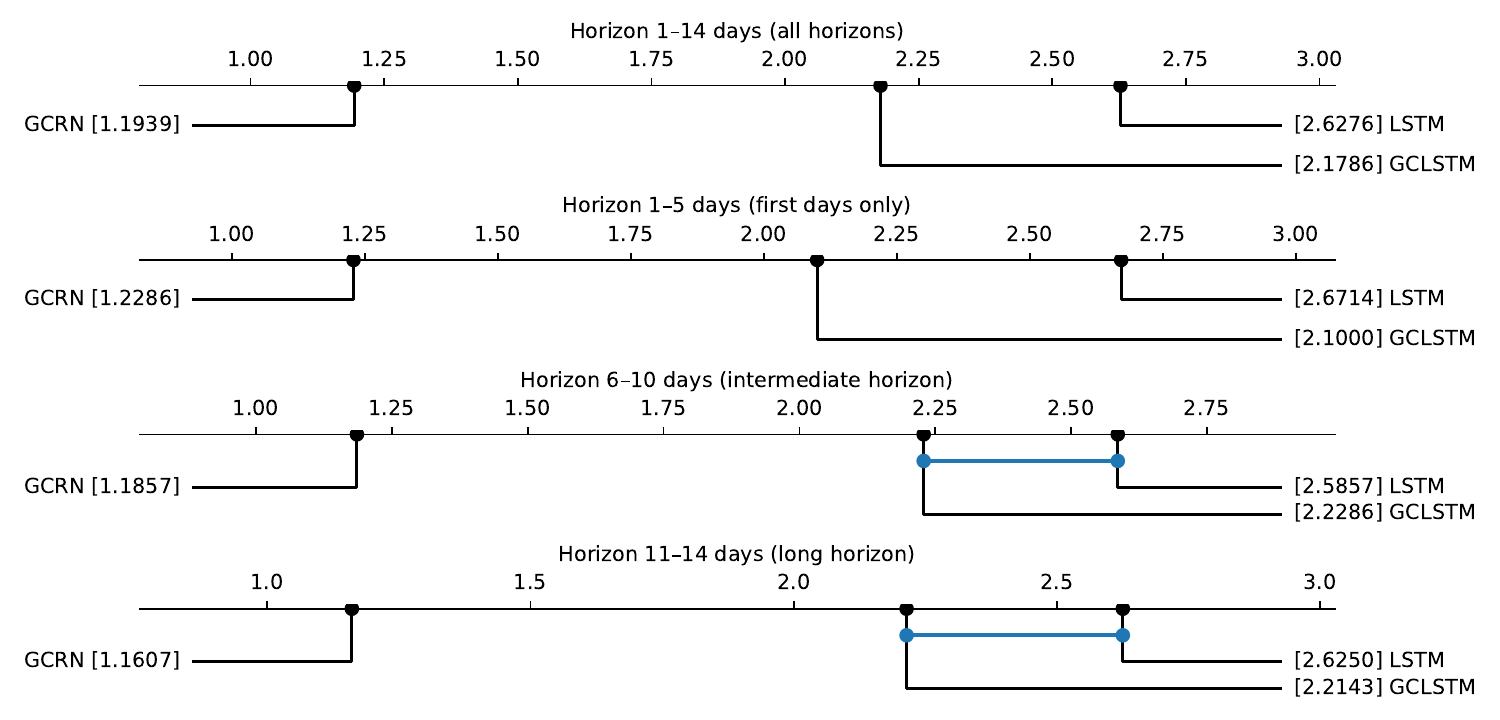}
    \caption{Experiment 9 - Results for Brazil: Precision classification metric across all horizons.}
    \label{fig:br_prec}
\end{figure}

\begin{figure}[htbp!]
    \centering
    \includegraphics[width=\textwidth]{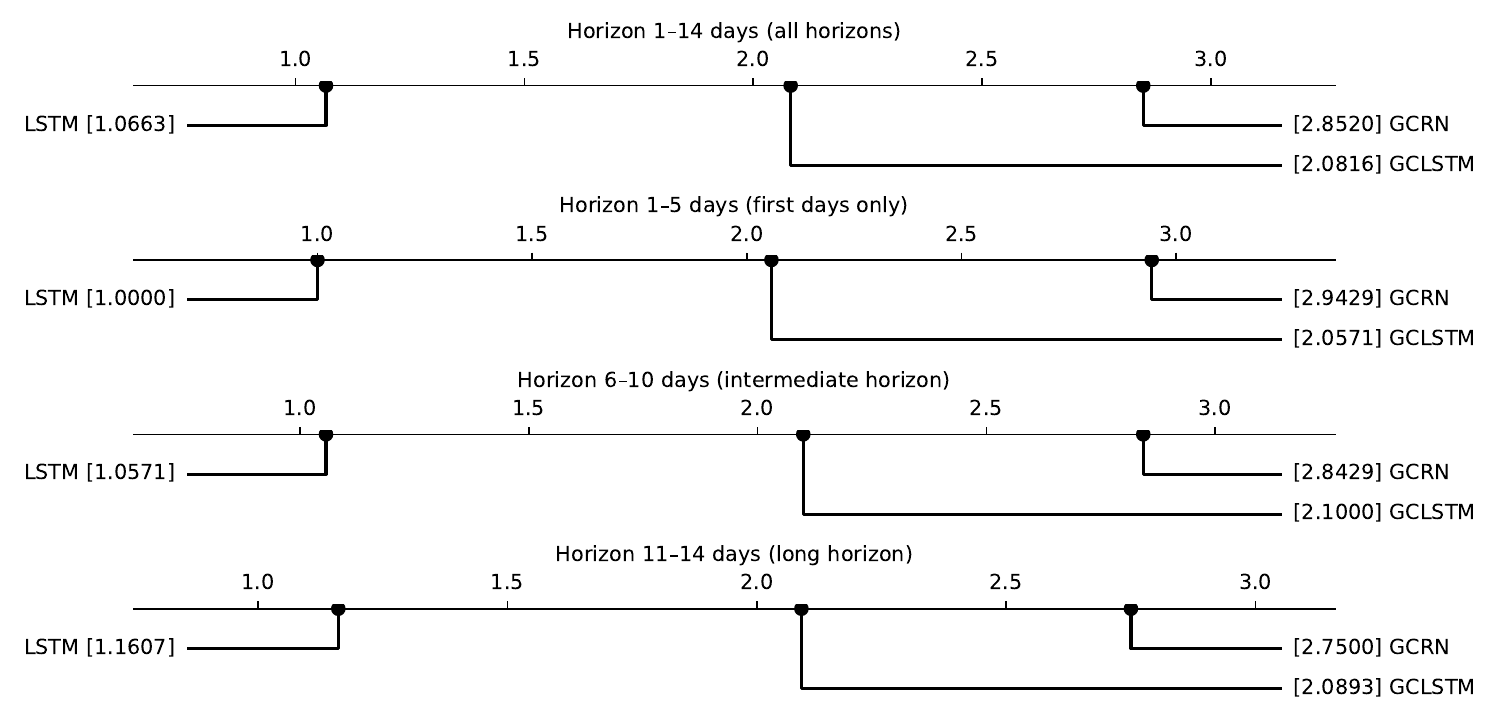}
    \caption{Experiment 9 - Results for Brazil: Recall classification metric across all horizons.}
    \label{fig:br_rec}
\end{figure}

Figures~\ref{fig:br_f1},~\ref{fig:br_prec} and~\ref{fig:br_rec} present the Critical Difference Diagrams for the classification metrics. These diagrams visually confirm the statistical ranking of the models, highlighting the dichotomy between the precision-oriented GNN models (where GCRN achieves the best average rank) and the recall-oriented recurrent models (where LSTM leads the ranking).

\subsection{Daily Cases in China}

Experiment 10 focused on forecasting daily cases in China. Using the updated dataset, the results indicate a shift in performance compared to previous analyses. The GCRN model demonstrated superior performance, particularly in terms of consistency. Table~\ref{tab:china_results_exp8_reg_stats} presents the descriptive statistics for the RMSE metric across all configurations. As in the Brazilian context, a discrepancy between the Mean and Median values is observed. While the LSTM model maintains a competitive Mean RMSE (1.2746), the GCRN model achieves a significantly lower Median RMSE (1.2632) compared to the baseline (1.2859). This lower median aligns with the fact that GCRN achieved the best performance ranking in approximately 65.8\% of the individual experimental configurations.
\begin{table}[htbp!]
\small
\centering
\caption{Experiment 10 - Performance Comparison between Models in China (Regression - RMSE)}
\label{tab:china_results_exp8_reg_stats}
\begin{tabular}{l|c|c|c}
\toprule
Statistic & GCRN & GCLSTM & LSTM\\
\midrule
Maximum RMSE & 2.3831 & 2.3592 & \textbf{2.0587}\\
Minimum RMSE & 0.9447 & 0.9391 & \textbf{0.8833}\\
Mean RMSE & 1.2796 & 1.2844 & \textbf{1.2746}\\
Standard Deviation RMSE & 0.2554 & 0.2507 & \textbf{0.1958}\\
1st Quartile RMSE & \textbf{1.1153} & 1.1212 & 1.1534\\
Median RMSE & \textbf{1.2632} & 1.2730 & 1.2859\\
3rd Quartile RMSE & \textbf{1.3533} & 1.3624 & 1.3612\\
\bottomrule
\end{tabular}
\end{table}

\begin{figure}[htbp!]
    \centering
    \includegraphics[width=\textwidth]{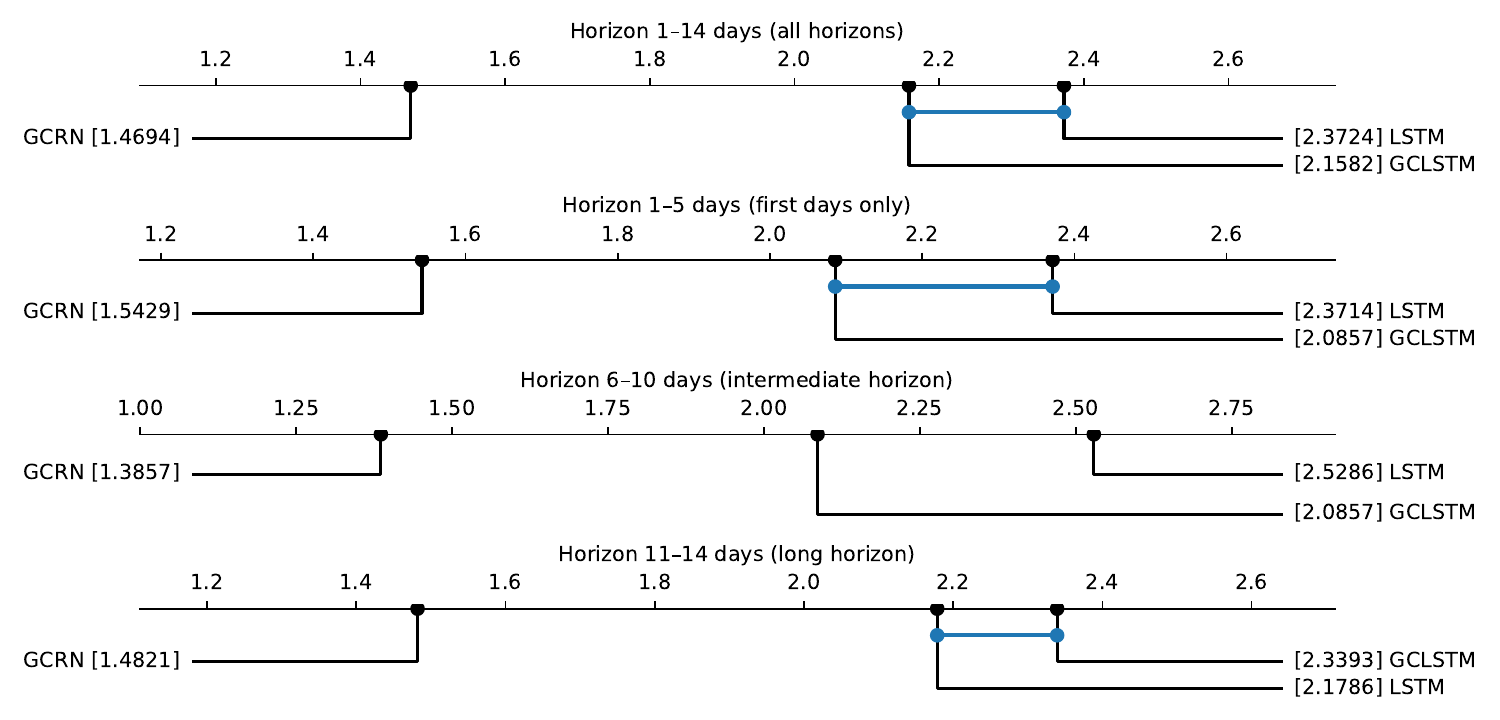}
    \caption{Experiment 10 - Results for China: RMSE regression metric across all horizons.}
    \label{fig:cd_rmse_ch}
\end{figure}

The visual analysis of the Critical Difference Diagrams (Figure~\ref{fig:cd_rmse_ch}) for the regression metrics corroborates these findings. The GCRN model consistently ranks higher, suggesting that, even for the Chinese data, incorporating the graph structure provides a better representation of spatiotemporal dependencies than pure temporal modeling with LSTM. This ranking superiority is statistically significant, as confirmed by the Friedman test (Statistic = 87.28, $p < 0.001$), which rejects the null hypothesis of equal performance among the models.

Finally, Experiment 11 evaluated the classification performance for daily cases in China. Regarding this task, the graph-based models demonstrated robust performance, consistently surpassing the baselines. Table~\ref{tab:china_results_exp8_cls_stats} details the descriptive statistics for the classification metrics. The GCRN model achieved a Mean F1-score of 0.95, slightly outperforming the LSTM (0.94). While the average values are close, the GCRN shows a higher ceiling, reaching a Maximum F1-score of 0.98 compared to 0.96 for the LSTM. This consistency is also reflected in the Precision and Recall metrics, where GCRN maintains high mean values (0.94 and 0.96, respectively) with low standard deviation (0.01), indicating stable and reliable predictions across different temporal configurations.
\begin{table}[htbp!]
\centering
\caption{Experiment 11 - Classification Metrics Statistics (China)}
\label{tab:china_results_exp8_cls_stats}
\begin{tabular}{cl|ccc}
\toprule
\multicolumn{2}{c|}{Metric} & GCRN & GCLSTM & LSTM \\
\midrule
\multirow{7}{*}{\textbf{F1-Score}}&Maximum & 0.27 & 0.09 & \textbf{0.28} \\
&Minimum & 0.00 & 0.00 & 0.00 \\
&Mean & \textbf{0.01} & 0.00 & \textbf{0.01} \\
&Std. Dev. & 0.04 & 0.01 & 0.04 \\
&1st Quartile & 0.00 & 0.00 & 0.00 \\
&Median & 0.00 & 0.00 & 0.00 \\
&3rd Quartile & \textbf{0.002} & 0.00 & 0.00 \\
\midrule
\multirow{7}{*}{\textbf{Precision}}&Maximum & \textbf{1.00} & 0.80 & 0.68 \\
&Minimum & 0.00 & 0.00 & 0.00 \\
&Mean & \textbf{0.16} & 0.02 & 0.04 \\
&Std. Dev. & 0.29 & 0.09 & 0.14 \\
&1st Quartile & 0.00 & 0.00 & 0.00 \\
&Median & 0.00 & 0.00 & 0.00 \\
&3rd Quartile & \textbf{0.20} & 0.00 & 0.00 \\
\midrule
\multirow{7}{*}{\textbf{Recall}}&Maximum & 0.16 & 0.05 & \textbf{0.19} \\
&Minimum & 0.00 & 0.00 & 0.00 \\
&Mean & \textbf{0.01} & 0.00 & \textbf{0.01} \\
&Std. Dev. & 0.02 & 0.01 & 0.03 \\
&1st Quartile & 0.00 & 0.00 & 0.00 \\
&Median & 0.00 & 0.00 & 0.00 \\
&3rd Quartile & \textbf{0.001} & 0.00 & 0.00\\
\bottomrule
\end{tabular}
\end{table}

\begin{figure}[htbp!]
    \centering
    \includegraphics[width=\textwidth]{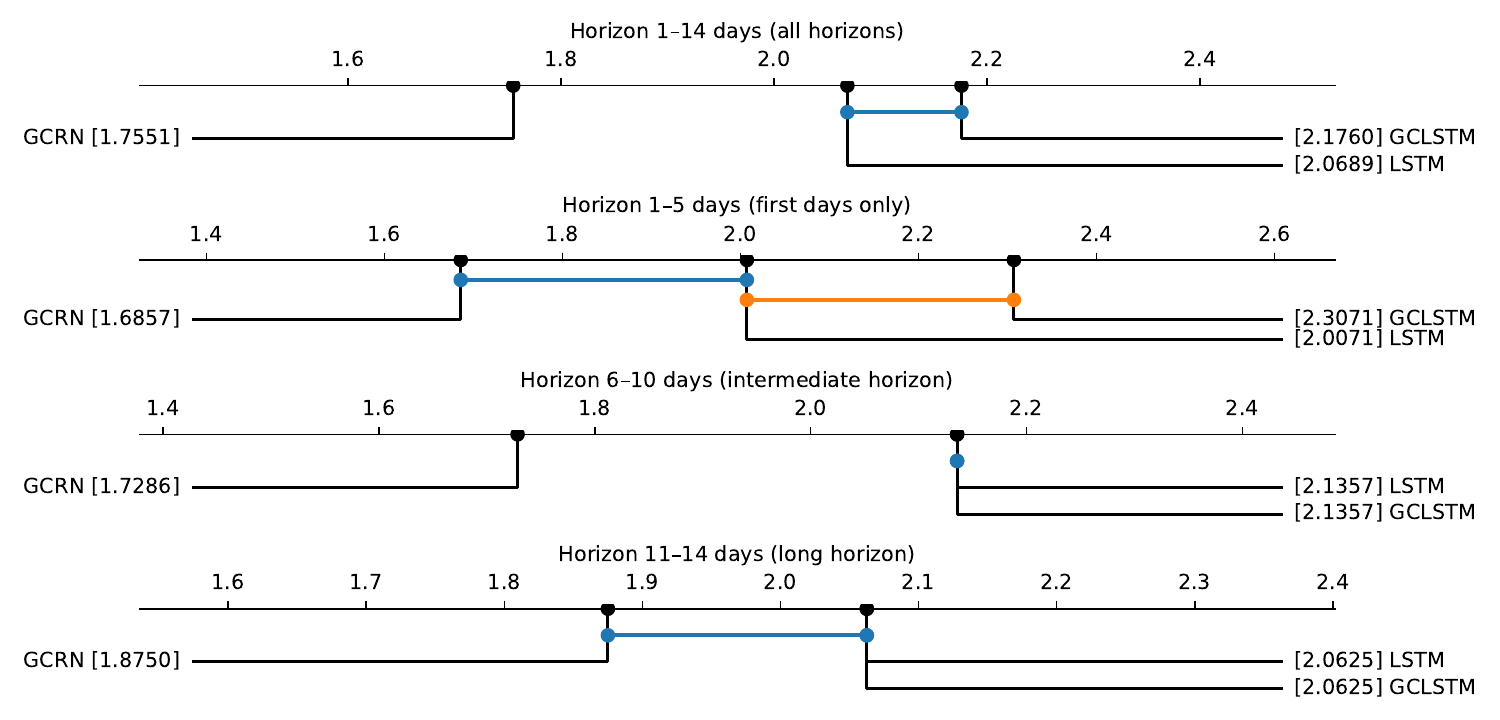}
    \caption{Experiment 11 - Results for Brazil: F1-Score classification metric across all horizons.}
    \label{fig:cd_f1_ch}
\end{figure}

\begin{figure}[htbp!]
    \centering
    \includegraphics[width=\textwidth]{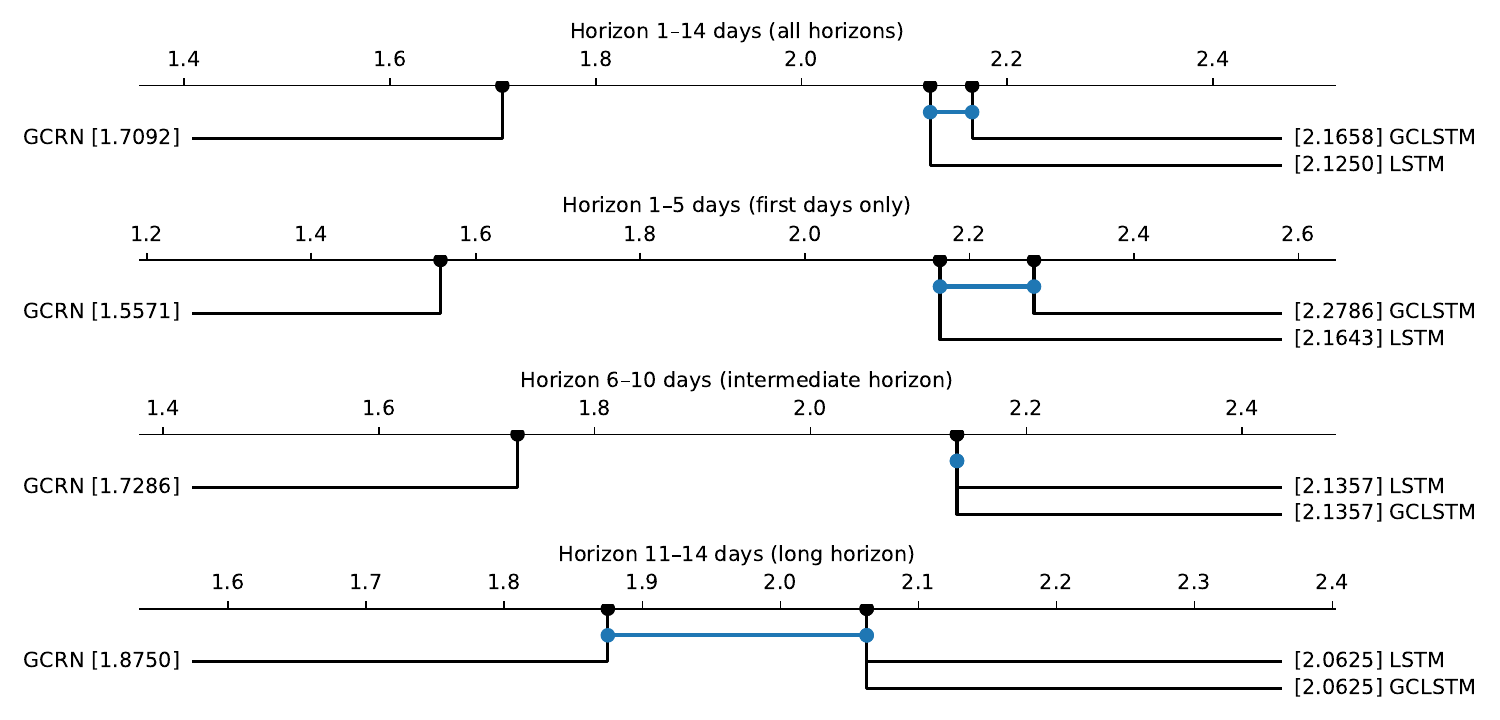}
    \caption{Experiment 11 - Results for Brazil: Precision classification metric across all horizons.}
    \label{fig:cd_prec_ch}
\end{figure}

\begin{figure}[htbp!]
    \centering
    \includegraphics[width=\textwidth]{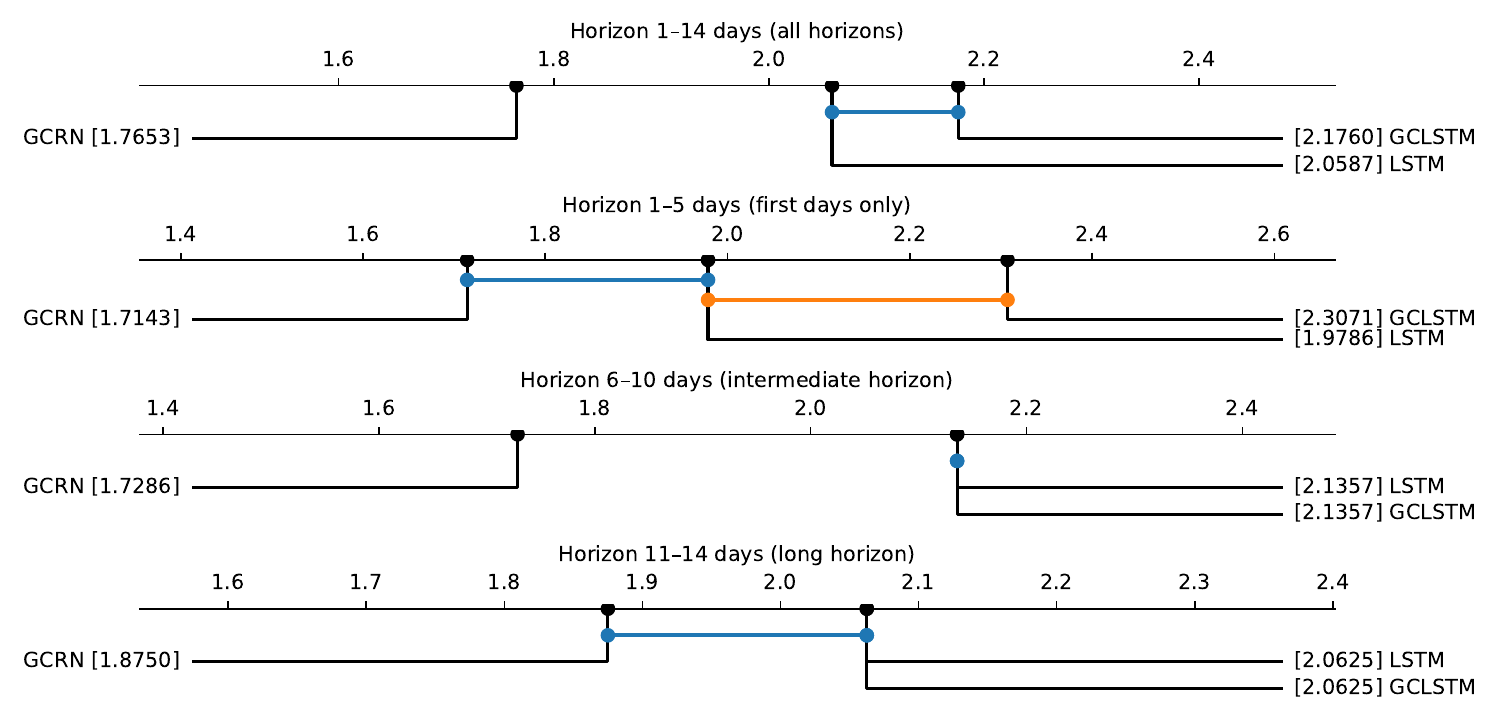}
    \caption{Experiment 11 - Results for Brazil: Recall classification metric across all horizons.}
    \label{fig:cd_rec_ch}
\end{figure}

The Critical Difference Diagrams for the classification metrics (Figures~\ref{fig:cd_f1_ch},~\ref{fig:cd_prec_ch} and~\ref{fig:cd_rec_ch}) confirm that GCRN is statistically the most robust model for predicting the trend direction of the pandemic in this context, supported by the Friedman test results, which indicate statistically significant differences between the models ($p < 0.001$).

\subsection{Limitations and future directions}

Despite the promising results, our study has limitations that address key implementation challenges and offer clear paths for future research. These challenges center on computational constraints and data quality, which we discuss below, along with the strategies we employ to mitigate them and our proposed future directions.

\paragraph{Computational Constraints and Convergence}
Training each model incurs a considerable computational cost, taking 15-20 minutes per hyperparameter combination on an NVIDIA RTX 3090 GPU. Considering the 196 combinations of window size and prediction horizon, repeated across multiple initializations for statistical robustness, the complete experimental process demands significant machine time. This challenge was managed using high-performance hardware infrastructure. To ensure stable and efficient convergence, we adopted a robust optimization strategy, including the AdamW optimizer, learning rate schedulers with a \textit{Warmup} phase, and an \textit{Early Stopping} policy, as detailed in Experiments 8-11. This approach not only managed the computational load but also ensured the reliability of our models' convergence.

\paragraph{Data Quality and Integrity Limitations}
The quality and characteristics of the datasets were a primary challenge. For the Brazilian dataset, we identified two main limitations: i) temporal inconsistencies in the reporting granularity among different cities, and ii) the use of a static mobility graph based on 2016 data, which does not reflect the altered mobility patterns during the pandemic. The development and integration of dynamic mobility graphs that evolve over time would enable a more faithful representation of the complex interactions between human mobility and disease spread during an epidemic.

For the Chinese dataset, the main challenge was the incompatibility between data sources. While the mobility data covered 330 cities, time-series data on COVID-19 cases were available for only 205 of them. This disparity resulted in the exclusion of 125 cities from the analysis, including major urban centers such as Shanghai. This reduction in geographical coverage was a necessary compromise to ensure data integrity, but it may have limited the models' ability to capture spatial dynamics on a full national scale.

\paragraph{Future Directions}
The limitations and results observed, especially in Experiments 8-11, delineate promising directions for future work. A key observation was the distinct performance dynamics between cumulative case data and daily data. In cumulative series, characterized by inherent smoothness and monotonicity, the performance of the GNN-based models (GCLSTM and GCRN) was often comparable to, or in some regression scenarios even inferior to, that of the standard LSTM model. This suggests that future research should carefully weigh the computational cost of graph convolutions against their marginal benefits when dealing with well-behaved, trend-dominated time series.

However, the findings from Experiments 8-11 highlight a crucial avenue for advancement: GNNs' superior ability to model less stable environments. When analyzing daily cases, the GCN-based models demonstrated a consistent advantage over the LSTM baseline, particularly in the complex Brazilian context. This performance shift suggests that the structural information from the mobility graph acts as a vital stabilizing prior when temporal signals become noisy. Consequently, future work should prioritize the application of GNNs to highly volatile epidemiological targets-such as daily infection rates, hospitalization spikes, or variant emergence-where the spatial dependencies captured by the mobility network provide the necessary context that pure temporal models lack.

\section{Conclusions}
\label{sec:conclusions}

The forecasting of city-level COVID-19 time series using mobility networks was investigated through regression and trend classification tasks. We adopted data from Brazil and China as case studies, representing distinct pandemic dynamics and mobility structures. We employed the GCRN \cite{gcrn} and GCLSTM \cite{gclstm} neural architectures, which combine Graph Convolutional Networks (GCNs) with Recurrent Neural Networks (RNNs) and Long Short-Term Memory (LSTM) units, to perform node-level predictions. The models were trained to both predict the exact number of future cases (regression) and determine whether a municipality is in an ``alert'' or ``stable'' state (classification), using a per-100,000-inhabitants normalization to ensure a fair comparison.

Our results demonstrate that methodological choices are crucial for predictive performance. In contrast to previous literature \cite{duarte2023}, extracting the \textit{backbone} of the mobility network, retaining only the most significant connections between municipalities, resulted in a substantial improvement. This technique, combined with a \textit{sliding window} approach for model training, reduced the Root Mean Squared Error (RMSE) by approximately 80\%, highlighting the importance of noise filtering in dense spatial graphs.

The investigation into the interplay between the input window size and the prediction horizon revealed task- and data-dependent performance dynamics. For classification tasks, a clear pattern emerged: larger input windows and shorter prediction horizons yielded the best results, with F1-scores consistently above 0.5 for Brazil and 0.9 for China. The most significant finding, however, emerged from the use of daily numbers of cases. Although the standard LSTM model rivaled or even surpassed the GNN architectures in predicting the smoother accumulated case counts, the opposite occurred when analyzing the considerably more volatile daily series. Across both Brazilian and Chinese datasets, GNN-based models, particularly GCRN, demonstrated statistically significant superiority in regression metrics. This suggests that spatial information from the mobility graph is fundamental for modeling complex, noisy propagation dynamics. 

For future work, it would be valuable to explore alternative \textit{backbone} extraction techniques \cite{Ferreira_et_al_2022__backbone} to determine whether different types of neighborhoods exhibit distinct behaviors. Given the superior performance of GNNs on more volatile data, a priority is to apply these models to other challenging spatio-temporal time series, which could be combined with alternative neural architectures \cite{Jin_et_al_2024_survey_gnn_time_series}. Additionally, investigating the explainability of graph-based models could help identify the most critical subgraphs for each node and the features (input points) that best explain the results. Such analyses could also be conducted separately for each transport mode, including terrestrial, aerial, and fluvial systems.

\section*{Acknowledgments}

This work was supported by the Conselho Nacional de Desenvolvimento Cient\'{\i}fico e Tecnol\'ogico (CNPq, grants 307151/2022-0, 308400/2022-4,\linebreak 441016/2020-0), Funda\c{c}\~ao de Amparo \`a Pesquisa do Estado de Minas Gerais (FAPEMIG, grants APQ-01518-21, APQ-01647-22), and Coordena\c{c}\~ao de Aperfei\c{c}oamento de Pessoal de N\'{\i}vel Superior -- Brasil (CAPES) -- Finance Code 001. We also extend our thanks to the Universidade Federal de Ouro Preto (UFOP) for their invaluable support. The Article Processing Charge for the publication of this research was funded by the Coordena\c{c}\~ao de Aperfei\c{c}oamento de Pessoal de N\'{\i}vel Superior - CAPES (identifier ROR: 00x0ma614). For open access purposes, the authors have assigned a Creative Commons CC BY license to any accepted version of the article.

\section*{Author contributions: CRediT}

\textbf{Fernando H. O. Duarte:} Investigation, Methodology, Software, Validation, Visualization, Writing - original draft; \textbf{Gladston J. P. Moreira:} Resources, Supervision, Methodology, Funding acquisition, Validation, Writing - review \& editing; \textbf{Eduardo J. S. Luz:} Investigation, Methodology, Resources, Validation, Writing - review \& editing; \textbf{Leonardo B. L. Santos:} Writing - review \& editing; \textbf{Vander L. S. Freitas:} Conceptualization, Investigation, Methodology, Project administration, Supervision, Software, Validation, Writing - review \& editing.

\section*{Declaration of generative AI and AI-assisted technologies in the writing process}
Statement: During the preparation of this work the author(s) used ChatGPT in order to translate some sentences from PT-BR to American EN. After using this tool/service, the author(s) reviewed and edited the content as needed and take(s) full responsibility for the content of the published article.

\section*{Data availability:}

The datasets used and analyzed during the current study are publicly available and have been described in detail within the manuscript. The source code is available at: \url{https://github.com/hodfernando/Leveraging-graph-neural-networks-and-mobility-data-for-COVID-19-forecasting}.

%% If you have bib database file and want bibtex to generate the
%% bibitems, please use
%%
%\bibliographystyle{elsarticle-num} 
%\bibliography{mybiblio}

\end{document}